\documentclass[11pt]{article}

\usepackage[preprint]{acl}

\usepackage{DejaVuSans}

\usepackage{pgfplots}
\pgfplotsset{compat=1.18}

\usepackage{inconsolata}

\usepackage[table]{xcolor}
\usepackage{booktabs}
\usepackage{multirow}
\usepackage{makecell}
\usepackage{colortbl}
\usepackage{xcolor}
\usepackage{booktabs}
\usepackage{amssymb}

\usepackage{tcolorbox}
\usepackage{times}
\usepackage{latexsym}
\usepackage[T1]{fontenc}
\usepackage[utf8]{inputenc}
\usepackage{microtype}
\usepackage{hyperref}
\usepackage{url}
\usepackage{booktabs}
\usepackage{xspace}
\usepackage{graphicx}
\usepackage{textcomp}
\usepackage{multicol}
\usepackage{multirow}
\usepackage{enumitem}
\usepackage{pifont}
\usepackage{subfigure}
\usepackage{float}
\usepackage{tikz}
\usepackage{subcaption}
\usepackage{lineno}
\usepackage{makecell}
\usepackage{CJKutf8}
\usepackage{inconsolata}
\usepackage{adjustbox}
\usepackage{placeins}

\usepackage{amsmath,amsfonts,bm}

\def\1{\bm{1}}

\DeclareMathAlphabet{\mathsfit}{\encodingdefault}{\sfdefault}{m}{sl}
\SetMathAlphabet{\mathsfit}{bold}{\encodingdefault}{\sfdefault}{bx}{n}

\usepackage{listings}
\usepackage{xcolor}
\usepackage{float}      
\usepackage{caption}    

\floatstyle{plain}
\newfloat{lstfloat}{htbp}{lop}
\floatname{lstfloat}{Listing}

\definecolor{promptbg}{RGB}{248,248,248}
\definecolor{promptframe}{RGB}{200,200,200}
\definecolor{promptkey}{RGB}{60,60,140}

\lstdefinestyle{promptstyle}{
    backgroundcolor=\color{promptbg},
    frame=single,
    rulecolor=\color{promptframe},
    framesep=4pt,
    basicstyle=\ttfamily\small,
    breaklines=true,
    breakatwhitespace=false,
    columns=fullflexible,
    keepspaces=true,
    showstringspaces=false,
    captionpos=b,
    aboveskip=4pt,
    belowskip=4pt,
    numbers=none,
    xleftmargin=4pt,
    xrightmargin=4pt,
}
\usepackage{listings}

\definecolor{darkblue}{rgb}{0, 0, 0.5}
\hypersetup{colorlinks=true, citecolor=darkblue, linkcolor=darkblue, urlcolor=darkblue}

\definecolor{mycolor}{RGB}{33, 95, 154}
\definecolor{custom_red}{RGB}{228, 54, 54}

\title{CPR for LLMs: Critical-Point Routing \\ against Catastrophic Forgetting in Domain Adaptation
}

\author{
\textbf{Kwangmin Ki}\textsuperscript{1\dag} \quad
\textbf{Yunhun Nam}\textsuperscript{1} \quad
\textbf{Jongheon Jeong}\textsuperscript{1} \quad
\textbf{Jaehyung Kim}\textsuperscript{2}
\\
\textsuperscript{1}Korea University
\qquad
\textsuperscript{2}Yonsei University
\\
\texttt{\{kwangminki,yh0326,jonghj\}@korea.ac.kr}
\qquad
\texttt{jaehyungk@yonsei.ac.kr}
}

\begin{document}
\maketitle

\begingroup
\renewcommand{\thefootnote}{}
\footnotetext{\textsuperscript{\dag}Work done during an internship at Yonsei University.}
\endgroup

\begin{abstract}
Supervised fine-tuning (SFT) is the de facto standard for adapting large language models (LLMs) to target domains, but it often degrades the model's general capabilities, a phenomenon known as catastrophic forgetting. 
Existing approaches typically modify the SFT loss to mitigate forgetting, but inevitably operate along a domain-generality trade-off. 
In this work, we step outside this trade-off by decoupling the two capabilities at the model level: 
we keep the original base model for general capability, and selectively invoke the SFT expert only when domain-specific knowledge is required.
Specifically, we propose \textbf{CPR} (\textbf{C}ritical-\textbf{P}oint \textbf{R}outing), a token-level routing framework 
between a base model and its expert derivative,
based on critical tokens where the base fails but the expert succeeds.
We train a lightweight hierarchical router that estimates expert-call probability per token, and pair it with a tailored inference procedure combining momentum smoothing and threshold gating. 
Across diverse model-domain configurations, CPR achieves state-of-the-art performance across all settings, surpassing SFT expert by 1.4-5.5\% in domain performance while recovering its general-capability drop from 3.4-14.5\% to at most 0.5\%, with minimal overhead from invoking the expert on only one-third of tokens.\footnote{ \url{https://github.com/kwang-min-ki/CPR}.}
\end{abstract}

\begin{figure}[t]
\centering
\includegraphics[width=\columnwidth]{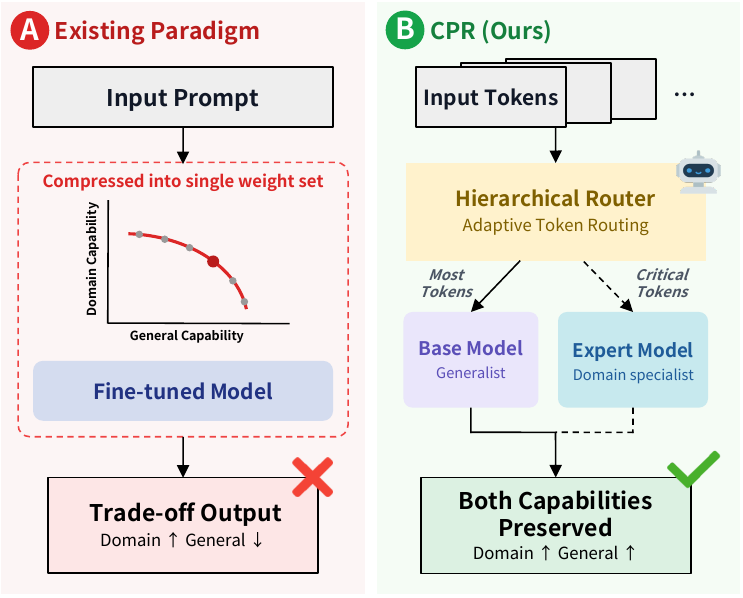} 
\caption{\textbf{Comparison of paradigms against catastrophic forgetting in SFT.} 
(A) Existing methods compress both capabilities into one model, yielding a trade-off. 
(B) CPR (Ours) decouples them by routing the expert only on critical tokens, preserving both.}
\label{fig:high}
\end{figure} 
\begin{figure*}[t]
\centering
\includegraphics[width=1.0\textwidth]{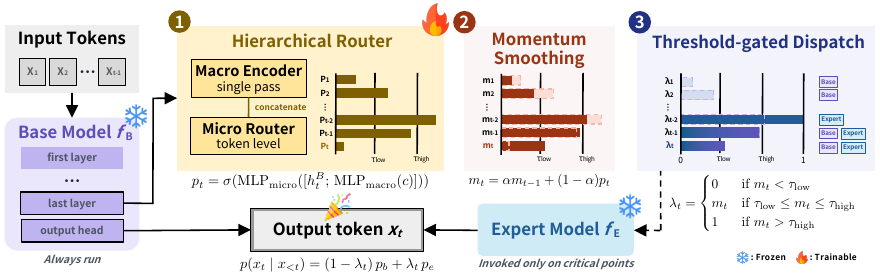} 
\caption{\textbf{Overview of CPR.} At each generation step $t$, (1) the base model $f_B$ produces a last-layer hidden state $h^B$ that feeds a hierarchical router $g_\phi$, composed of a query-level macro encoder and a token-level micro router. 
(2) The router's raw output $p_t$ is stabilized by momentum smoothing $m_t = \alpha·m_{t-1} + (1-\alpha)·p_t$, and then (3) mapped to a dispatch weight $\lambda_t$ via threshold-gated 3-way dispatch: hard base ($m_t < \tau_{low}$), soft blend ($\tau_{low} \leq m_t \leq \tau_{high}$), or hard expert ($m_t > \tau_{high}$). 
The final token distribution is a convex combination $p(x_t | x_{<t}) = (1-\lambda_t)·p_b + \lambda_t·p_e$.}
\label{fig:low}
\vspace{-1mm}
\end{figure*}

\section{Introduction} \label{sec1}
Large language models (LLMs) trained on broad corpora have become widely deployed across diverse applications, from open-ended dialogue to complex multi-step reasoning \citep{wei2022chain, kojima2022large}. 
To improve the performance on specific domains such as mathematics \citep{toshniwal2025openmathinstruct} and medicine \citep{han2023medalpaca}, supervised fine-tuning (SFT) on domain data has become the standard approach.
However, domain-specific SFT often degrades the model's general capabilities, a phenomenon known as catastrophic forgetting \citep{luo2025empirical, kotha2024understanding}. 
This degradation is problematic because domain tasks themselves build on general capabilities such as language understanding, instruction following, and multi-step reasoning; a model whose domain capability is raised at the expense of these foundations becomes brittle in practice \citep{liu2024more}.
The key challenge in domain adaptation is therefore to improve domain performance and preserve general capabilities simultaneously.

To address this, most existing approaches follow a regularization-based paradigm, modifying SFT loss to reduce forgetting by adding distribution preservation terms or reweighting token-level loss \citep{wu2025generalization, diao2026entropy, nam2026learning}.
Despite clear progress, these methods share a common limitation: they compress both domain and general capabilities into a single weight set (Figure \ref{fig:high}(a)), and thus remain bound to a domain-general trade-off that no existing method has fully eliminated \citep{lin2025sft}.\footnote{Our empirical results for this are presented in Figure \ref{fig:tradeoff}.} 

This motivates a fundamentally different question: 
\textit{rather than compressing both capabilities into a single model, can we decouple them at the model level and selectively use them?} 
Inter-model routing has been studied along two strands: (i) query-level cost-quality routing \citep{ong2024routellm, chen2023frugalgpt, ding2024hybrid} and (ii) fine-grained routing for cross-domain capability fusion \citep{feng2025don, xiong2026token, shen2024learning}. 
However, its use as an architectural remedy for catastrophic forgetting in domain-specific SFT remains largely underexplored.

In this work, we propose \textbf{CPR} (\textbf{C}ritical-\textbf{P}oint \textbf{R}outing), a new framework that preserves general capabilities after SFT, by collaborating an SFT-tuned expert model and an initial base model via token-level routing (Figure \ref{fig:high}(b)).
The expert is invoked only on the sparse set of tokens where it is genuinely needed, leaving the rest to the base, so that each token is served by whichever model is best suited to it.
Specifically, we first define \textit{critical points} as tokens where the base fails but the expert succeeds.
Then, we train a lightweight hierarchical router on them, composed of a query-level macro encoder and a token-level micro router operating on the base model's last-layer hidden states. 
At inference, since binary dispatch is destabilized by short-range noise and boundary ambiguity, we introduce two tailored mechanisms: (i) momentum-based smoothing to absorb short-term router noise, and (ii) threshold-gated 3-way dispatch to blend two distributions on ambiguous tokens.
Beyond stabilizing dispatch, these mechanisms also yield efficiency gains by avoiding the per-step dual-model cost. 
An overview of CPR is presented in Figure \ref{fig:low}.

We validate CPR on two representative LLMs, Gemma3-4B \citep{Kamath2025Gemma3T} and Llama3.1-8B \citep{grattafiori2024llama}, adapted to math \citep{cobbe2021training} and medical \citep{jin2019pubmedqa} domains, evaluated over six benchmarks spanning domain and general capabilities. 
CPR attains the highest overall average in all four model-domain settings, surpassing vanilla SFT in target domain performance while nearly fully recovering its general-capability drop. 
For example, with Gemma3-4B in the math domain, CPR surpasses vanilla SFT by 2.8\% in domain performance; at the same time, it improves upon SFT by 8.9\%, exceeding even the base model's performance by 2.2\%. 
More importantly, CPR consistently outperforms state-of-the-art regularization-based baselines \citep{wu2025generalization,diao2026entropy,nam2026learning} and routing-based baselines at coarser granularity \citep{ong2024routellm, feng2025don}, empirically demonstrating the superiority of the proposed token-level adaptive routing.
In addition, CPR matches or exceeds performance while invoking the expert on only \textasciitilde30\% of tokens, substantially reducing inference latency compared to conventional collaborative decoding methods.
We further show that CPR generalizes to publicly available external experts without additional SFT, demonstrating its robustness to how the expert is produced.
We hope that this work offers a new perspective on mitigating catastrophic forgetting in domain-specific LLM adaptation.
\section{Related Work}

\paragraph{Catastrophic Forgetting in Domain SFT.} 
Despite its effectiveness, domain-specific SFT is widely reported to degrade LLMs' general capabilities. 
\citet{luo2025empirical} empirically show that catastrophic forgetting arises consistently across 1B-7B LLMs under continual instruction tuning. 
\citet{kotha2024understanding} attribute this to fine-tuning skewing the model's implicit task inference toward the SFT distribution, thereby suppressing pre-existing capabilities. 
Furthermore, \citet{lobo2025impact} show that task-specific fine-tuning consistently reduces both the accuracy and the faithfulness of chain-of-thought reasoning, indicating that SFT can disturb the reasoning machinery that domain tasks rely on.

To mitigate this trade-off, prior work has concentrated on modifying the next token prediction training loss. 
DFT \citep{wu2025generalization} rescales the per-token loss by the model's token probability to neutralize an implicit inverse-probability reward; 
EAFT \citep{diao2026entropy} uses token-level entropy as a soft gate to suppress destructive gradients on confident-conflict tokens;
LfU \citep{nam2026learning} regularizes representations to remain consistent with those after an undesirable update, preserving prior knowledge.
As the core idea is to compress domain and general capabilities into a single weight set, they fundamentally operate along the trade-off.
In contrast, CPR separates two capabilities at the model level and then merges them via adaptive routing, escaping the single-weight trade-off.

\paragraph{Inter-model Routing for LLM Decoding.}
The proposed approach is situated within the broader paradigm of inter-model routing which dynamically combines multiple LLMs at inference.
Prior studies generally fall into two distinct categories. 
The first category routes an entire query to one of several models, driven by a cost-quality trade-off.
RouteLLM \citep{ong2024routellm} learns a router from human preferences to dispatch between a strong and a weak LLM; 
FrugalGPT \citep{chen2023frugalgpt} cascades LLMs under a budget constraint;
Hybrid LLM \citep{ding2024hybrid} routes queries based on difficulty and a tunable quality target. 
Although these methods differ from ours in both objective and granularity, they establish input-dependent inter-model dispatch as a core design principle.

The second category combines model outputs at finer granularity for cross-domain capability fusion. 
Co-LLM \citep{shen2024learning} models the per-token deferral as a latent variable trained via marginal likelihood; 
Switch Generation \citep{feng2025don} alternates between pretrained, finetuned, and aligned checkpoints at the patch level to recover skills lost in alignment;
FusionRoute \citep{xiong2026token} selects a token-level expert while adding a complementary logit from the router itself. 
Although they target fusing heterogeneous strengths rather than mitigating forgetting with SFT, these works motivate the token-level granularity of our approach, enabling more precise expert invocation.
\section{Method} \label{sec3}

\subsection{Problem Setup}

Let $f_B$ denote the original base model and $f_E$ the domain expert model
derived from $f_B$.
Given an input prompt followed by generation up to total length $L$
, we denote the full sequence as $x = (x_1, \ldots, x_{L_p}, \ldots, x_L)$, where $L_p$ is the index of the first generated token. The input prompt may consist of optional few-shot demonstrations followed by a question; we denote the question span as $\mathcal{S}_q = \{s_q, \ldots, L_p-1\}$, where $s_q$ is the index of the first question token (with $s_q = 1$ in the 0-shot case). 
At each generation step $t \in \{L_p, \ldots, L-1\}$, models $f_B$ and $f_E$ produce next-token distributions $p_b$ and $p_e$ over the vocabulary, conditioned on $x_{<t}$.
Then, we aim to produce a per-token output distribution as a convex combination of the two:
\begin{equation}
p(x_t \mid x_{<t}) = (1 - \lambda_t)\, p_b + \lambda_t\, p_e,
\end{equation}
where $\lambda_t \in [0, 1]$ is a token-level dispatch weight that 
takes one of three forms: $\lambda_t = 0$, $\lambda_t = 1$, or $\lambda_t \in [\tau_{low}, \tau_{high}]$. 
To determine $\lambda_t$, we (i) introduce a lightweight hierarchical router $g_\phi$, (ii) train it with token-level critical-point labels (Section \ref{sec:3.2}), and (iii) apply momentum smoothing and threshold gating at inference (Section \ref{sec:inference}).

{This convex combination can be viewed from a latent-variable perspective, where the router estimates whether the base or expert model should generate each token. 
Marginalizing over this choice naturally motivates a token-dependent mixture of their output distributions.
If the router output is interpreted as an estimate of the posterior probability that the expert should be selected, marginalizing over this latent choice yields exactly the convex mixture above. 
We therefore use its temporally smoothed estimate $m_t$ as the mixture weight $\lambda_t$ in the ambiguous Soft Blend regime.}

\subsection{Training}\label{sec:3.2}

\textbf{Router Architecture.}
The router $g_\phi$ is organized into two stages: a \textit{macro encoder} that summarizes the domain relevance of the entire question once per sequence, and a \textit{micro router} that makes a per-token decision at every generation step. 
These two components are implemented as $\mathrm{MLP}_{\mathrm{macro}}$ and $\mathrm{MLP}_{\mathrm{micro}}$ respectively, 
both of which are 2-layer ReLU MLPs, and we denote the full set of trainable parameters as $\phi$.

The macro encoder first forms a question context vector $c \in \mathbb{R}^d$ by mean-pooling the base model's last-layer hidden states over the question span $\mathcal{S}_q$, and maps it to a domain summary $z_{\mathrm{dom}} \in \mathbb{R}^{d_h}$:
\begin{equation}
c = \frac{1}{|\mathcal{S}_q|} \sum_{i \in \mathcal{S}_q} h_i^B,\quad z_{\mathrm{dom}} = \mathrm{MLP}_{\mathrm{macro}}(c),
\end{equation}
where $h_i^{B} \in \mathbb{R}^d$ is the base model's last-layer hidden state at position $i$, $d$ is the base model's hidden dimension, and $d_h$ is the router's hidden dimension.

At each generation step $t$, the micro router takes the current hidden state $h_t^{B}$ together with the macro summary $z_{\mathrm{dom}}$, and outputs the probability of invoking the expert $p_t$:
\begin{equation}
g_\phi(h_t^{B}, c) = \sigma(\mathrm{MLP}_{\mathrm{micro}}([h_t^{B};\, z_{\mathrm{dom}}])),
\end{equation}
\begin{equation}
p_t = g_\phi(h_t^{B}, c) \in (0, 1),
\end{equation}
where $[\cdot \,;\, \cdot]$ denotes vector concatenation along the feature dimension. 

\medskip
\noindent\textbf{Labeling.}
To train the router $g_\phi$, we need a token-level label indicating whether invoking the expert is necessary. 
We construct this label automatically via teacher-forcing comparison between the base and the expert models.
For each training sequence $x$ and each generation step $t \in \{L_p, \ldots, L-1\}$, we obtain the greedy predictions of both models:
\begin{equation}
\hat{x}_t^{(B)} = \arg\max_v p_b(v \mid x_{<t}),
\end{equation}
\begin{equation}
\hat{x}_t^{(E)} = \arg\max_v p_e(v \mid x_{<t}).
\end{equation}
Comparing these predictions against the ground-truth token $x_t$, we assign a token-level label $z_t \in \{0, 1, \varnothing\}$:
\begin{equation}
z_t =
\begin{cases}
0 & \text{if } \hat{x}_t^{(B)} = x_t \\
1 & \text{if } \hat{x}_t^{(B)} \neq x_t \;\wedge\; \hat{x}_t^{(E)} = x_t \\
\varnothing & \text{otherwise}
\end{cases}
\label{eq:labeling}
\end{equation}
The label reflects the necessity of an expert call rather than the correctness of the expert itself. 
Tokens that the base already predicts correctly are assigned $z_t = 0$ regardless of whether the expert also succeeds and tokens with $z_t = 1$ are referred to as \textit{critical points}.

\medskip
\noindent\textbf{Training of Router.}
We freeze both $f_B$ and $f_E$, and train only router parameters $\phi$ via binary cross-entropy (BCE) against the critical-point labels. 
Since critical points are typically sparse ($z_t = 0$ tokens far outnumber $z_t = 1$), we use a re-weighted BCE to mitigate class-imbalance:
\begin{equation}
\mathcal{L}(\phi) = \frac{1}{|\mathcal{V}|} \sum_{t \in \mathcal{V}} \ell_t,
\end{equation}
\begin{equation}
\ell_t = -\bigl[ w^{+} z_t \log p_t + (1 - z_t)\log(1 - p_t) \bigr],
\end{equation}
where $\mathcal{V} = \{t : z_t \neq \varnothing\}$ is the set of valid supervision tokens. The positive weight is set to $w^{+} = \sqrt{N_0 / N_1}$, where $N_0$ and $N_1$ are the corpus-level counts of $z_t = 0$ and $z_t = 1$ tokens, respectively. 
The square-root form moderates the over-weighting of critical points that naive inverse-frequency reweighting would induce.

\subsection{Inference}
\label{sec:inference}
Using the trained router with a hard binary decision at each step is undesirable for two reasons: (i) raw non-deterministic router output $p_t$ may oscillate between adjacent tokens or flip on ambiguous boundaries, destabilizing dispatch quality \citep{ding2024hybrid}, and (ii) invoking the expert at every step incurs unnecessary per-step latency. We address both with \textit{momentum-based smoothing} and \textit{threshold-gated 3-way dispatch}.

\medskip
\noindent\textbf{Momentum-based Smoothing.} Rather than using the raw $p_t$, we maintain its exponential moving average $m_t$ to absorb short-range noise:
\begin{equation}
m_0 = 0, \quad m_t = \alpha m_{t-1} + (1 - \alpha) p_t,
\label{eq:momentum}
\end{equation}
where $\alpha \in [0, 1)$ is the momentum decay factor. We use $\alpha = 0.5$ as the default; larger $\alpha$ increases the inertia of past decisions.

\medskip
\noindent\textbf{Threshold-gated 3-way Dispatch.} Based on $m_t$, we introduce two thresholds 
$\tau_{\mathrm{low}}, \tau_{\mathrm{high}}$ with $0 < \tau_{\mathrm{low}} < \tau_{\mathrm{high}} < 1$ 
which partition the momentum value $m_t$ into three regimes that determine $\lambda_t$:
\begin{equation}
\lambda_t = \begin{cases}
0   & \text{if } m_t < \tau_{\mathrm{low}}, \\
m_t & \text{if } \tau_{\mathrm{low}} \leq m_t \leq \tau_{\mathrm{high}}, \\
1   & \text{if } m_t > \tau_{\mathrm{high}}.
\end{cases}
\label{eq:dispatch}
\end{equation}
We use $\tau_{\mathrm{low}} = 0.35$ and $\tau_{\mathrm{high}} = 0.65$ by default. 

The three regimes correspond to distinct dispatch behaviors: 
\emph{Hard Base} regime ($\lambda_t = 0$) and \emph{Hard Expert} regime ($\lambda_t = 1$) deterministically draw from $p_b$ and $p_e$ respectively.
In the intermediate \emph{Soft Blend} regime ($\lambda_t = m_t$), the routing decision is ambiguous; $m_t$ itself serves as the dispatch weight, yielding a convex combination of $p_b$ and $p_e$ that softly interpolates between the two models.

\medskip
\noindent\textbf{Selective Expert Invocation.} 
The 3-way dispatch yields an inference-efficiency benefit beyond routing quality. 
Because the base hidden state $h_t^{B}$ feeds the router, the base model is always run; the expert, however, is only executed in the Soft Blend and Hard Expert regimes. 
Thus, unlike collaborative decoding approaches such as ensembling and contrastive decoding that execute both models at every step, CPR reduces the number of sequential expert forward passes by handling skipped positions in a single batched KV cache catch-up pass. 
For instance, suppose the expert remains inactive for seven consecutive decoding steps and is invoked at the next step. 
It first synchronizes its KV cache for the seven skipped positions in a single batched catch-up pass and then computes the current step. 
If it remains active, subsequent tokens use ordinary per-step expert decoding with the synchronized cache. 
Thus, skipped positions do not require separate sequential expert passes. 
Although this catch-up can make the total FLOPs comparable to those of collaborative decoding, reducing sequential expert passes substantially lowers wall-clock latency because autoregressive LLM decoding is largely memory-bandwidth-bound.

\begin{table*}[t]
\renewcommand{\arraystretch}{1.0}
\centering
\scriptsize
\setlength{\tabcolsep}{4pt}
\resizebox{\textwidth}{!}{%
\begin{tabular}{lccccccccc}
\toprule

\multicolumn{1}{c}{} 
& \multicolumn{4}{c}{\textbf{Math Domain}} 
& \multicolumn{4}{c}{\textbf{General Domain}} 
& \multirow{2}{*}{\textbf{Overall}} \\

\cmidrule(lr){2-5} \cmidrule(lr){6-9}

\textbf{Method}
& \textbf{GSM8K\textsuperscript{\(\star\)}}
& \textbf{ASDiv}
& \textbf{SVAMP}
& \textbf{Average}
& \textbf{MMLU}
& \textbf{CSQA}
& \textbf{ARC-C}
& \textbf{Average}
& \textbf{Average} \\

\midrule

\multicolumn{10}{c}{\textit{Gemma3-4B}} \\
\midrule

Base           
& 29.11 & 44.51 & 48.67 
& 40.76 
& 54.16 & 59.13 & 64.42 
& 59.24 
& 50.00 \\

SFT (Expert)   
& 51.25 & 58.00 & 50.67 
& 53.31 {\color{blue}(+12.55)} 
& 49.17 & 52.91 & 55.38 
& 52.49 {\color{red}(-6.75)} 
& 52.90 {\color{blue}(+2.90)} \\

DFT                 
& 32.84 & 48.55 & 42.33 
& 41.24 {\color{gray}(+0.48)} 
& 45.95 & 57.66 & 47.59 
& 50.40 {\color{red}(-8.84)} 
& 45.82 {\color{red}(-4.18)}\\

EAFT                 
& 50.04 & 61.92 & 57.67 
& \textbf{56.54} {\color{blue}(+15.78)} 
& 51.15 & 48.87 & 54.10 
& 51.37 {\color{red}(-7.87)} 
& 53.95 {\color{blue}(+3.95)} \\

LfU                  
& 37.39 & 53.67 & 43.33 
& 44.80 {\color{blue}(+4.04)} 
& 50.41 & 53.56 & 53.94 
& 52.64 {\color{red}(-6.60)} 
& 48.72 {\color{red}(-1.28)} \\

\cmidrule(lr){1-10}

Switch Generation    
& 46.93 & 57.43 & 49.67 
& 51.34 {\color{blue}(+10.58)} 
& 50.90 & 52.90 & 57.25 
& 53.68 {\color{red}(-5.56)} 
& 52.51 {\color{blue}(+2.51)} \\

RouteLLM             
& 49.63 & 57.87 & 50.67 
& 52.72 {\color{blue}(+11.96)} 
& 49.91 & 52.99 & 56.01 
& 52.97 {\color{red}(-6.27)} 
& 52.85 {\color{blue}(+2.85)} \\

Ensemble             
& 53.08 & 60.38 & 54.00 
& 55.82 {\color{blue}(+15.06)} 
& 52.98 & 59.95 & 61.95 
& 58.29 {\color{gray}(-0.95)} 
& \underline{57.06} {\color{blue}(+7.06)} \\

Contrastive Decoding 
& 45.41 & 50.28 & 52.00 
& 49.23 {\color{blue}(+8.47)} 
& 54.63 & 59.71 & 65.87 
& \underline{60.07} {\color{gray}(+0.83)} 
& 54.65 {\color{blue}(+4.65)} \\

\cellcolor{blue!10}CPR \textbf{(Ours)} 
& \cellcolor{blue!10} 49.58 
& \cellcolor{blue!10} 62.60 
& \cellcolor{blue!10} 56.00 
& \cellcolor{blue!10} \underline{56.06} {\color{blue}(+15.30)} 
& \cellcolor{blue!10} 55.89 
& \cellcolor{blue!10} 61.10 
& \cellcolor{blue!10} 67.24 
& \cellcolor{blue!10} \textbf{61.41} {\color{blue}(+2.17)} 
& \cellcolor{blue!10} \textbf{58.74} {\color{blue}(+8.74)} \\

\midrule

\multicolumn{10}{c}{\textit{Llama3.1-8B}} \\
\midrule

Base           
& 34.72 & 35.84 & 44.33 
& 38.30 
& 57.23 & 61.18 & 71.67 
& \textbf{63.36} 
& 50.83 \\

SFT (Expert)       
& 45.87 & 58.79 & 60.00 
& 54.89 {\color{blue}(+16.59)} 
& 44.57 & 52.74 & 49.40 
& 48.90 {\color{red}(-14.46)} 
& 51.90 {\color{blue}(+1.07)} \\

DFT                 
& 41.67 & 57.91 & 59.87 
& 53.15 {\color{blue}(+14.85)} 
& 47.90 & 60.52 & 46.28 
& 51.57 {\color{red}(-11.79)} 
& 52.36 {\color{blue}(+1.53)}\\

EAFT                 
& 47.23 & 62.43 & 63.00 
& 57.55 {\color{blue}(+19.25)} 
& 43.20 & 53.07 & 45.17 
& 47.15 {\color{red}(-16.21)} 
& 52.35 {\color{blue}(+1.52)} \\

LfU                  
& 56.56 & 61.87 & 58.67 
& 59.03 {\color{blue}(+20.73)} 
& 56.36 & 65.50 & 67.49 
& \underline{63.12} {\color{gray}(-0.24)} 
& \underline{61.08} {\color{blue}(+10.25)} \\

\cmidrule(lr){1-10}

Switch Generation    
& 46.10 & 58.27 & 60.33 
& 54.90 {\color{blue}(+16.60)} 
& 47.47 & 52.98 & 51.01 
& 50.49 {\color{red}(-12.87)} 
& 52.69 {\color{blue}(+1.86)} \\

RouteLLM             
& 45.72 & 59.56 & 61.00 
& 55.43 {\color{blue}(+17.13)} 
& 45.43 & 53.33 & 49.23 
& 49.33 {\color{red}(-14.03)} 
& 52.38 {\color{blue}(+1.55)} \\

Ensemble             
& 51.30 & 62.80 & 64.00 
& \underline{59.37} {\color{blue}(+21.07)} 
& 52.39 & 60.14 & 56.71 
& 56.41 {\color{red}(-6.95)} 
& 57.89 {\color{blue}(+7.06)} \\

Contrastive Decoding 
& 49.60 & 60.89 & 61.33 
& 57.27 {\color{blue}(+18.97)} 
& 51.61 & 59.28 & 60.05 
& 56.98 {\color{red}(-6.38)} 
& 57.13 {\color{blue}(+6.30)} \\

\cellcolor{blue!10}CPR \textbf{(Ours)} 
& \cellcolor{blue!10} 51.80 
& \cellcolor{blue!10} 62.70 
& \cellcolor{blue!10} 66.67 
& \cellcolor{blue!10} \textbf{60.39} {\color{blue}(+22.09)} 
& \cellcolor{blue!10} 58.97 
& \cellcolor{blue!10} 64.26 
& \cellcolor{blue!10} 65.25 
& \cellcolor{blue!10} 62.83 {\color{gray}(-0.53)} 
& \cellcolor{blue!10} \textbf{61.61} {\color{blue}(+10.78)} \\

\bottomrule
\end{tabular}%
}
\caption{\textbf{Main results on the math domain.}  Test accuracy (\%) on in-domain math benchmarks and out-of-domain general benchmarks. \textsuperscript{\(\star\)} marks the SFT training source and parentheses show change (\%) relative to the base model. Overall Average is the mean of the math and general averages. The best and second best scores are highlighted in \textbf{bold} and \underline{underline}.}
\label{table:main_math}
\end{table*}
\begin{table*}[t]
\renewcommand{\arraystretch}{1.0}
\centering
\scriptsize
\setlength{\tabcolsep}{4pt}
\resizebox{\textwidth}{!}{%
\begin{tabular}{lccccccccc}
\toprule

\multicolumn{1}{c}{} 
& \multicolumn{4}{c}{\textbf{Medical Domain}} 
& \multicolumn{4}{c}{\textbf{General Domain}} 
& \multirow{2}{*}{\textbf{Overall}} \\

\cmidrule(lr){2-5} \cmidrule(lr){6-9}

\textbf{Method}
& \textbf{PubMedQA\textsuperscript{\(\star\)}}
& \textbf{MedQA}
& \textbf{CareQA}
& \textbf{Average}
& \textbf{MMLU}
& \textbf{CSQA}
& \textbf{ARC-C}
& \textbf{Average}
& \textbf{Average} \\

\midrule

\multicolumn{10}{c}{\textit{Gemma3-4B}} \\
\midrule

Base           
& 54.30 & 27.97 & 38.13 & 40.13 
& 54.16 & 59.13 & 64.42 & 59.24 
& 49.69 \\

SFT (Expert)        
& 75.70 & 41.08 & 39.29 & 52.02 {\color{blue}(+11.89)} 
& 51.89 & 56.67 & 58.96 & 55.84 {\color{red}(-3.40)} 
& 53.93 {\color{blue}(+4.24)} \\

DFT                 
& 56.50 & 38.10 & 40.39 & 45.00 {\color{blue}(+4.87)}
& 50.41 & 59.38 & 61.36 & 57.05 {\color{red}(-2.19)} 
& 51.03 {\color{blue}(+1.34)} \\

EAFT                 
& 73.80 & 34.60 & 37.24 & 48.55 {\color{blue}(+8.42)} 
& 45.39 & 55.22 & 56.71 & 52.44 {\color{red}(-6.80)} 
& 50.49 {\color{gray}(+0.80)} \\

LfU                  
& 66.10 & 34.25 & 39.20 & 46.52 {\color{blue}(+6.39)} 
& 49.92 & 57.00 & 62.95 & 56.62 {\color{red}(-2.62)} 
& 51.57 {\color{blue}(+1.88)} \\

\cmidrule(lr){1-10}

Switch Generation    
& 75.80 & 40.61 & 40.09 & 52.17 {\color{blue}(+12.04)} 
& 51.97 & 57.27 & 58.79 & 56.01 {\color{red}(-3.23)} 
& 54.09 {\color{blue}(+4.40)} \\

RouteLLM             
& 59.58 & 30.77 & 38.72 & 43.02 {\color{blue}(+2.89)} 
& 53.04 & 60.85 & 64.59 & 59.49 {\color{gray}(+0.25)} 
& 51.26 {\color{blue}(+1.57)} \\

Ensemble             
& 71.20 & 46.95 & 43.39 & \textbf{53.85} {\color{blue}(+13.72)} 
& 56.12 & 60.03 & 67.67 & \underline{61.27} {\color{blue}(+2.03)} 
& \underline{57.56} {\color{blue}(+7.87)} \\

Contrastive Decoding 
& 67.30 & 45.09 & 39.82 & 50.74 {\color{blue}(+10.61)} 
& 55.39 & 58.31 & 68.77 & 60.82 {\color{blue}(+1.58)} 
& 55.78 {\color{blue}(+6.09)} \\

\cellcolor{blue!10}CPR \textbf{(Ours)}  
& \cellcolor{blue!10} 66.80 
& \cellcolor{blue!10} 48.20 
& \cellcolor{blue!10} 45.19 
& \cellcolor{blue!10} \underline{53.40} {\color{blue}(+13.27)} 
& \cellcolor{blue!10} 54.82 
& \cellcolor{blue!10} 61.26 
& \cellcolor{blue!10} 69.67 
& \cellcolor{blue!10}\textbf{61.92} {\color{blue}(+2.68)} 
& \cellcolor{blue!10} \textbf{57.66} {\color{blue}(+7.97)} \\

\midrule

\multicolumn{10}{c}{\textit{Llama3.1-8B}} \\
\midrule

Base           
& 68.00 & 43.99 & 50.84 & 54.28 
& 57.23 & 61.18 & 71.67 & 63.36
& 58.82 \\

SFT (Expert)  
& 75.20 & 52.00 & 50.40 & 59.20 {\color{blue}(+4.92)}
& 53.31 & 59.38 & 57.94 & 56.88 {\color{red}(-6.48)}
& 58.04 {\color{gray}(-0.78)} \\

DFT                 
& 75.00 & 47.27 & 45.73 & 56.00 {\color{blue}(+1.72)}
& 54.23 & 64.13 & 64.72 & 61.03 {\color{red}(-2.33)}
& 58.52 {\color{gray}(-0.30)} \\

EAFT                 
& 75.50 & 52.70 & 50.82 & 59.67 {\color{blue}(+5.39)}
& 56.30 & 63.43 & 64.02 & 61.25 {\color{red}(-2.11)}
& 60.46 {\color{blue}(+1.64)} \\

LfU                  
& 75.60 & 53.05 & 52.60 & 60.42 {\color{blue}(+6.14)}
& 59.04 & 66.31 & 69.26 & \textbf{64.87} {\color{blue}(+1.51)}
& 62.65 {\color{blue}(+3.83)} \\

\cmidrule(lr){1-10}

Switch Generation    
& 75.30 & 51.37 & 49.60 & 58.76 {\color{blue}(+4.48)}
& 54.80 & 59.85 & 59.01 & 57.89 {\color{red}(-5.47)}
& 58.32 {\color{gray}(-0.50)} \\

RouteLLM             
& 70.20 & 43.83 & 52.36 & 55.46 {\color{blue}(+1.18)}
& 57.55 & 60.02 & 70.82 & 62.80 {\color{gray}(-0.56)}
& 59.13 {\color{gray}(+0.31)} \\

Ensemble             
& 76.00 & 54.22 & 53.98 & \underline{61.40} {\color{blue}(+7.12)}
& 57.06 & 66.55 & 61.16 & 61.59 {\color{red}(-1.77)}
& 61.50 {\color{blue}(+2.68)} \\

Contrastive Decoding 
& 74.80 & 53.15 & 55.78 & 61.24 {\color{blue}(+6.96)}
& 58.35 & 65.98 & 68.85 & \underline{64.39} {\color{blue}(+1.03)}
& \underline{62.82} {\color{blue}(+4.00)} \\

\cellcolor{blue!10}CPR \textbf{(Ours)} 
& \cellcolor{blue!10}74.60 
& \cellcolor{blue!10}53.86 
& \cellcolor{blue!10}56.62 
& \cellcolor{blue!10} \textbf{61.69} {\color{blue}(+7.41)}
& \cellcolor{blue!10}58.89 
& \cellcolor{blue!10}68.93 
& \cellcolor{blue!10}65.27
& \cellcolor{blue!10}64.36 {\color{blue}(+1.00)}
& \cellcolor{blue!10} \textbf{63.03} {\color{blue}(+4.21)} \\

\bottomrule
\end{tabular}%
}
\caption{\textbf{Main results on the medical domain.} Test accuracy (\%) on in-domain medical benchmarks and out-of-domain general 
benchmarks. \textsuperscript{\(\star\)} marks the SFT training source and parentheses show change (\%) relative to the base model. Overall Average is the mean of the medical and general averages. The best and second best 
scores are highlighted in \textbf{bold} and \underline{underline}.}
\label{table:main_medical}
\end{table*}

\section{Experiments} \label{sec4}

We evaluate CPR across math and medical domains and analyze its components, robustness to training-data scale, inference efficiency, routing behavior, and generalization to external experts.
Additional evaluations in Appendix~\ref{app:quantitative} include comparisons with intra-model routing baselines, experiments on finance and open-ended instruction following, and a lightweight LoRA expert configuration.

\subsection{Setups}

\textbf{Models.} We evaluate CPR on two representative open-source LLMs: \textit{Gemma3-4B} \citep{Kamath2025Gemma3T} and \textit{Llama3.1-8B} \citep{grattafiori2024llama}. 
For each base model, we construct a domain expert by vanilla SFT on two domains, math and medical, yielding four model-domain combinations in total. 
The math expert is trained on the GSM8K training set (approximately 8K examples), while the medical expert is trained on a size-matched sample drawn from the PubMedQA artificial split. 
To prevent label-distribution bias, we sample the medical training set with a balanced split between the yes and no classes (4,000 examples each). 
Math training data are formatted as 0-shot chain-of-thought (CoT), and medical data are formatted as 2-shot CoT with answer options randomly shuffled to prevent positional bias; 2-shot demonstrations are drawn from MMLU.

\medskip
\noindent\textbf{Benchmarks.} 
Each setting is evaluated on six benchmarks covering both domain and general capabilities. 
The general capability is measured uniformly across all settings as the average over MMLU \citep{hendrycks2020measuring}, CommonsenseQA \citep{talmor2019commonsenseqa}, and ARC-C \citep{clark2018think}. 
Domain capability is measured as the average over GSM8K \citep{cobbe2021training}, ASDiv \citep{miao2020diverse}, and SVAMP \citep{patel2021nlp} for the math setting, and over PubMedQA \citep{jin2019pubmedqa}, MedQA \citep{jin2021disease}, and CareQA \citep{arias2025automatic} for the medical setting. 
For the evaluation, we adopt CoT prompting;  
following the same protocol as training, math benchmarks use 0-shot CoT, while medical and general benchmarks use 2-shot CoT with MMLU CoT demonstrations. 
We report the overall average as the arithmetic mean of the general and domain averages, providing a single metric for how well the two capabilities are jointly attained.

\medskip
\noindent\textbf{Baselines.} 
We consider two groups of state-of-the-art baselines. 
The first group consists of single-model baselines that integrate both capabilities into one weight set. 
This group first includes the base and SFT expert model, which define the two extremes that any routing scheme operates between, corresponding to the references for general and domain capability respectively. 
Next, regularization-based methods that modify the SFT loss to mitigate forgetting are considered: DFT \citep{wu2025generalization}, EAFT \citep{diao2026entropy}, and LfU \citep{nam2026learning}.
The second group consists of multi-model collaboration baselines that route or combine the base and the expert. 
Within this group, routing-based methods dispatch at different granularities: query-level \citep{ong2024routellm} and patch-level \citep{feng2025don}.
We used the officially released checkpoints rather than re-training on our SFT data, as the latter underperforms in our preliminary experiments (see Appendix~\ref{app:retrained-routers}).
Decoding-based methods mix the two output distributions at every decoding step with static weights, always invoking both models: Ensemble and Contrastive Decoding \citep{li2023contrastive}. 
All baselines share the same base model and SFT expert model.
Full implementation details including training and inference hyperparameters are deferred to Appendix \ref{app:implementation}.

\subsection{Main Results}

The results are presented in Tables \ref{table:main_math} and \ref{table:main_medical}. 
Here, CPR attains the highest overall average across all model-domain combinations, improving over the base model by 4.2\% to 10.8\%.
It recovers nearly all of the general-capability drop incurred by SFT expert model while simultaneously surpassing the expert in domain performance. 
For instance, on Llama3.1-8B math, the SFT expert loses 14.5\% on general capability, whereas CPR loses only 0.5\%; on the domain side, the expert improves by 16.6\% over the base, while CPR improves by 22.1\%, an additional 5.5\% gain over the expert.

At the same time, regularization-based methods are observed to exhibit a clear domain-general trade-off. 
Across four settings, they reduce forgetting compared to SFT expert slightly yet still incur up to 16.2\% general-capability drop, whereas CPR is the only method that improves both axes simultaneously, indicating that the trade-off is intrinsic to single-weight regularization and can be removed by decoupling two capabilities at the model level.

In addition, we observe that existing routing methods, RouteLLM (query-level) and Switch Generation (patch-level), underperform CPR.
On the math domain, both improve the overall average by 1.6-2.9\% over the base, whereas CPR improves it by 8.7-10.8\%. 
Since all routing baselines share the same base model and the same SFT expert, this controlled comparison empirically establishes that the proposed token-level routing is a more effective remedy for catastrophic forgetting than coarser-granularity alternatives.

Lastly, while Ensemble and Contrastive Decoding are effective by doubling per-token compute, CPR is more effective and efficient (see Section \ref{sec4.3.efficiency} for efficiency results). 
In the math domain, CPR outperforms both baselines by 1.7-4.5\% in overall average across the two backbones.
Similarly, in the medical domain, CPR achieves the highest overall average in every setting.

\begin{table}[t]
\renewcommand{\arraystretch}{1.0}
\centering
\setlength{\tabcolsep}{6pt}
\resizebox{\columnwidth}{!}{%
\begin{tabular}{lccc}
\toprule
\textbf{Method}
& \textbf{Math Average}
& \textbf{General Average}
& \textbf{Overall Average} \\
\midrule
Base
& 39.53 & 61.30 & 50.42 \\
\midrule
\multicolumn{4}{c}{\textit{Router Architecture}} \\
\midrule
w/o Macro
& 57.52 & 60.67 & 59.10 \\

w/o Micro
& 49.63 & 61.28 & 55.45 \\

Hierarchical \textbf{(Default)}
& 58.23 & 62.12 & \textbf{60.18} \\
\midrule
\multicolumn{4}{c}{\textit{Momentum Decay $\alpha$}} \\
\midrule
$\alpha$ = 0.0
& 55.97 & 60.55 & 58.26 \\
$\alpha$ = 0.5 \textbf{(Default)}
& 58.23 & 62.12 & \textbf{60.18} \\
$\alpha$ = 0.9
& 46.32 & 62.71 & 54.52 \\
\midrule
\multicolumn{4}{c}{\textit{Dispatch Thresholds $(\tau_{\mathrm{low}}, \tau_{\mathrm{high}})$}} \\
\midrule
(0.5, 0.5)
& 51.98 & 62.35 & 57.16 \\
(0.35, 0.65) \textbf{(Default)}
& 58.23 & 62.12 & \textbf{60.18} \\
(0.0, 1.0)
& 58.07 & 62.26 & 60.17 \\
\bottomrule
\end{tabular}%
}
\caption{\textbf{Ablation studies of CPR.}
Test accuracy (\%) averaged over Gemma3-4B and Llama3.1-8B on in-domain math and out-of-domain general benchmarks, under variations in (a) router architecture, (b) momentum decay $\alpha$, and (c) thresholds $(\tau_{low}, \tau_{high})$. The best score is highlighted in \textbf{bold}.}
\label{table:ablation_mini}
\end{table}

\subsection{Analyses} 

\noindent\textbf{Ablation Study.}
\label{sec4.3.ablation}
To understand how each component contributes to CPR, we ablate (a) router architecture, (b) momentum decay factor $\alpha$, and (c) dispatch thresholds $(\tau_{\text{low}}, \tau_{\text{high}})$, with results summarized in Table \ref{table:ablation_mini} (See full benchmark numbers in Appendix \ref{app:ablation}). 
(a) Both components of the hierarchical router contribute complementarily. Removing the macro encoder degrades overall accuracy with 1.5\% drop on the general domain, indicating that the query-level signal acts as a domain prior that suppresses spurious expert calls on out-of-domain queries. Removing the micro router causes an 8.6\% drop on the math domain, showing that token-level dispatch is essential for precisely identifying critical points where the expert is needed.
(b) For the momentum factor in Eq.\ref{eq:momentum}, where larger $\alpha$ puts more weight on past decisions, the default $\alpha$ = 0.5 consistently performs best. $\alpha$ = 0.0 ignores previous history and causes dispatch to oscillate near the boundary, while $\alpha$ = 0.9 over-relies on previous history and delays switching into the expert at the onset of critical regions, leading to a large domain drop with 11.9\% on math average. The asymmetry suggests that over-reliance on past decisions is more harmful than ignoring them when critical points are sparse and abrupt.
(c) For the thresholds in Eq.\ref{eq:dispatch}, collapsing the gate to hard switching at $(0.5, 0.5)$ removes the Soft Blend regime and drops domain accuracy with 6.3\% drop on math average, whereas expanding to $(0.0, 1.0)$ attains comparable accuracy on both axes but always invokes the expert that leads to increased latency.
Overall, the three components play complementary roles, and the default configuration generalizes without per-setting tuning.


\begin{table}[t]
\centering
\small
\setlength{\tabcolsep}{3pt}
\begin{tabular}{lccc}

\toprule
Method & GSM8K & MMLU & Average \\
\midrule
Base
& 29.11 & 54.16 & 41.64 \\
\midrule
SFT (1K)
& 36.32 \textcolor{blue}{(+7.21)}
& 53.91 \textcolor{gray}{(-0.25)}
& 45.12 \textcolor{blue}{(+3.48)} \\

CPR (1K)
& 39.73 \textcolor{blue}{(+10.62)}
& 55.64 \textcolor{blue}{(+1.48)}
& \textbf{47.69} \textcolor{blue}{(+6.05)} \\
\midrule
SFT (2K)
& 38.44 \textcolor{blue}{(+9.33)}
& 49.56 \textcolor{red}{(-4.60)}
& 44.00 \textcolor{blue}{(+2.36)} \\

CPR (2K)
& 41.61 \textcolor{blue}{(+12.50)}
& 54.76 \textcolor{gray}{(+0.60)}
& \textbf{48.19} \textcolor{blue}{(+6.55)} \\
\midrule
SFT (4K)
& 45.41 \textcolor{blue}{(+16.30)}
& 52.32 \textcolor{red}{(-1.84)}
& 48.87 \textcolor{blue}{(+7.23)} \\

CPR (4K)
& 45.26 \textcolor{blue}{(+16.15)}
& 55.73 \textcolor{blue}{(+1.57)}
& \textbf{50.50} \textcolor{blue}{(+8.86)} \\
\midrule
SFT (8K)
& 51.25 \textcolor{blue}{(+22.14)}
& 49.17 \textcolor{red}{(-4.99)}
& 50.21 \textcolor{blue}{(+8.58)} \\

CPR (8K)
& 49.58 \textcolor{blue}{(+20.47)}
& 55.89 \textcolor{blue}{(+1.73)}
& \textbf{52.74} \textcolor{blue}{(+11.11)} \\
\bottomrule
\end{tabular}
\caption{\textbf{Robustness to training-data scale.}
Results with Gemma3-4B under 1K, 2K, 4K, and 8K GSM8K training subsets, evaluated on GSM8K and MMLU.}
\label{tab:data_scale}
\end{table}
\begin{figure}[t]
\centering
\includegraphics[width=\columnwidth]{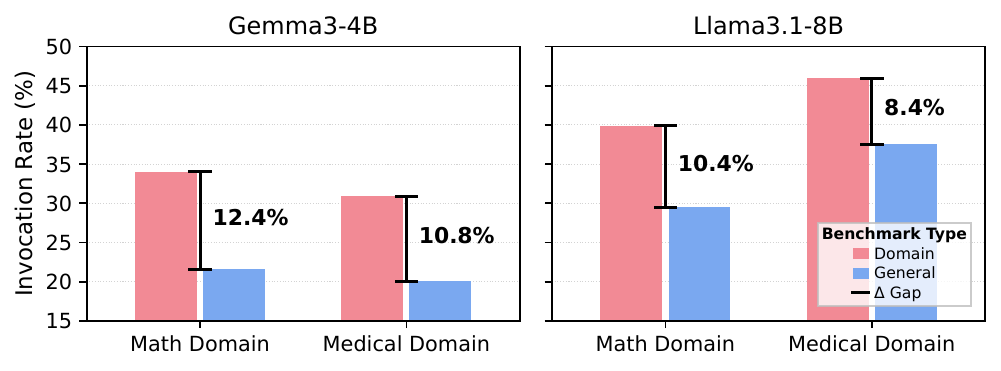} 
\caption{\textbf{Expert invocation rate (\%).} Average percentage of critical tokens routed to the expert on in-domain vs. out-of-domain general benchmarks, with $\Delta$ gap line indicating the difference between them.}
\label{fig:efficiency}
\end{figure}

\begin{table}[t]
\centering
\small
\setlength{\tabcolsep}{5pt}
\begin{tabular}{lcc}
\toprule
Method & Latency (s) & Tokens/sec \\
\midrule
\multicolumn{3}{c}{\textit{Gemma3-4B}} \\
\midrule
Expert
    & 5.581 (1.00$\times$)
    & 17.4 \\
Ensemble
    & 10.548 (1.89$\times$)
    & 9.4 \\
CPR \textbf{(Ours)}
    & 7.826 (1.40$\times$)
    & 13.4 \\
\midrule
\multicolumn{3}{c}{\textit{Llama3.1-8B}} \\
\midrule
Expert
    & 2.286 (1.00$\times$)
    & 37.2 \\
Ensemble
    & 5.147 (2.25$\times$)
    & 19.0 \\
CPR \textbf{(Ours)}
    & 3.302 (1.44$\times$)
    & 25.5 \\
\bottomrule
\end{tabular}
\caption{\textbf{Decoding runtime comparison.}
 Latency and throughput on GSM8K with Gemma3-4B and Llama3.1-8B, measured on 100 samples per method.
}
\label{table:runtime}
\end{table}
\begin{table*}[t]
\centering
\scriptsize
\setlength{\tabcolsep}{4pt}
\resizebox{\textwidth}{!}{%
\begin{tabular}{lccccccccc}
\toprule

\multicolumn{1}{c}{} 
& \multicolumn{4}{c}{\textbf{Math Domain}} 
& \multicolumn{4}{c}{\textbf{General Domain}} 
& \multirow{2}{*}{\textbf{Overall}} \\

\cmidrule(lr){2-5} \cmidrule(lr){6-9}

\textbf{Method}
& \textbf{GSM8K}
& \textbf{ASDiv}
& \textbf{SVAMP}
& \textbf{Average}
& \textbf{MMLU}
& \textbf{CSQA}
& \textbf{ARC-C}
& \textbf{Average}
& \textbf{Average} \\
\midrule

\multicolumn{10}{c}{\textit{Llama3.1-8B (OpenMath2-Llama3.1-8B)}} \\
\midrule

Base           
& 34.72 & 35.84 & 44.33 
& 38.30 
& 57.23 & 61.18 & 71.67 
& 63.36 
& 50.83 \\

Expert        
& 85.67 & 83.38 & 87.00 
& \textbf{85.35} {\color{blue}(+47.05)}
& 38.49 & 49.22 & 47.78 
& 45.16 {\color{red}(-18.20)}
& 65.26 {\color{blue}(+14.43)}\\

\cellcolor{blue!10}CPR \textbf{(Ours)}        
& \cellcolor{blue!10}70.76 & \cellcolor{blue!10}72.93 & \cellcolor{blue!10}75.33 
& \cellcolor{blue!10}73.00 {\color{blue}(+34.70)}
& \cellcolor{blue!10}56.14 & \cellcolor{blue!10}68.22 & \cellcolor{blue!10}67.92
& \cellcolor{blue!10}\textbf{64.09} {\color{gray}(+0.73)}
& \cellcolor{blue!10}\textbf{68.55} {\color{blue}(+17.72)}\\

\midrule

\multicolumn{10}{c}{\textit{Qwen2.5-1.5B (Qwen2.5-Math-1.5B)}} \\
\midrule

Base           
& 64.06 & 76.44 & 74.33 
& 71.61 
& 53.20 & 69.62 & 68.26 
& \textbf{63.69} 
& 67.65 \\

Expert        
& 71.80 & 81.26 & 85.00
& \textbf{79.35} {\color{blue}(+7.74)}
& 40.15 & 31.45 & 49.49
& 40.36 {\color{red}(-23.33)}
& 59.86 {\color{red}(-7.79)}\\

\cellcolor{blue!10}CPR \textbf{(Ours)}        
& \cellcolor{blue!10}68.89 & \cellcolor{blue!10}78.70 & \cellcolor{blue!10}79.67 
& \cellcolor{blue!10}75.75 {\color{blue}(+4.14)}
& \cellcolor{blue!10}53.10 & \cellcolor{blue!10}68.30 & \cellcolor{blue!10}67.94 
& \cellcolor{blue!10}63.11 {\color{gray}(-0.58)}
& \cellcolor{blue!10}\textbf{69.43} {\color{blue}(+1.78)}\\

\bottomrule
\end{tabular}%
}
\caption{\textbf{Generalization to external experts.} CPR paired with external math experts instead of our SFT experts on Llama3.1-8B and Qwen2.5-1.5B. Math benchmarks for Qwen use 2-shot CoT due to limited 0-shot capacity; all other settings are identical to the main experiments. The best score is highlighted in \textbf{bold}.}\label{table:ext_expert}
\end{table*}

\medskip\noindent\textbf{Robustness to Training Data Scale.}
We examine CPR's robustness as the amount of domain training data decreases.
Using Gemma3-4B, we train both the expert and the router on 1K, 2K, 4K, and 8K subsets of GSM8K, where smaller subsets jointly weaken the expert and reduce the amount of router supervision.
As shown in Table~\ref{tab:data_scale}, CPR remains effective across all data scales.
Even with only 1K examples, where the SFT expert improves GSM8K by 7.21\% over the base, CPR improves domain accuracy by 10.62\% while maintaining MMLU above the base level.
Across all scales, CPR achieves the highest overall average and keeps MMLU at or above the base performance.

\medskip\noindent\textbf{Inference Efficiency.} \label{sec4.3.efficiency}
Another goal of CPR is to minimize the latency overhead inherent to token-level routing. 
Figure \ref{fig:efficiency} visualizes the fraction of tokens routed to the expert on domain and general benchmarks across both math and medical domains.
The expert is invoked on only about one-third of tokens across overall benchmarks, confirming that critical points are sparse. More importantly, a consistent domain vs. general gap is observed across all four settings, showing that the router concentrates expert calls where they are actually needed rather than thresholding noise. 
This selectivity is also the mechanism by which CPR preserves general capability: on general queries, the router defaults to the base for most tokens, sidestepping the distributional shift introduced by the expert.

Table~\ref{table:runtime} reports measured wall-clock latency and throughput on GSM8K for both Gemma3-4B and Llama3.1-8B.
For Gemma3-4B, CPR incurs 1.40$\times$ latency relative to expert-only decoding, substantially below Ensemble at 1.89$\times$.
The same advantage holds for Llama3.1-8B despite its higher expert invocation rate, with CPR at 1.44$\times$ compared with 2.25$\times$ for Ensemble.
These measurements include the full KV-cache catch-up cost described in Section~\ref{sec:inference}.

\begin{figure}[t]
\centering
\begin{minipage}{\columnwidth}
\tiny\textbf{Query 1}
\end{minipage}\\[1pt]
\includegraphics[width=\columnwidth]{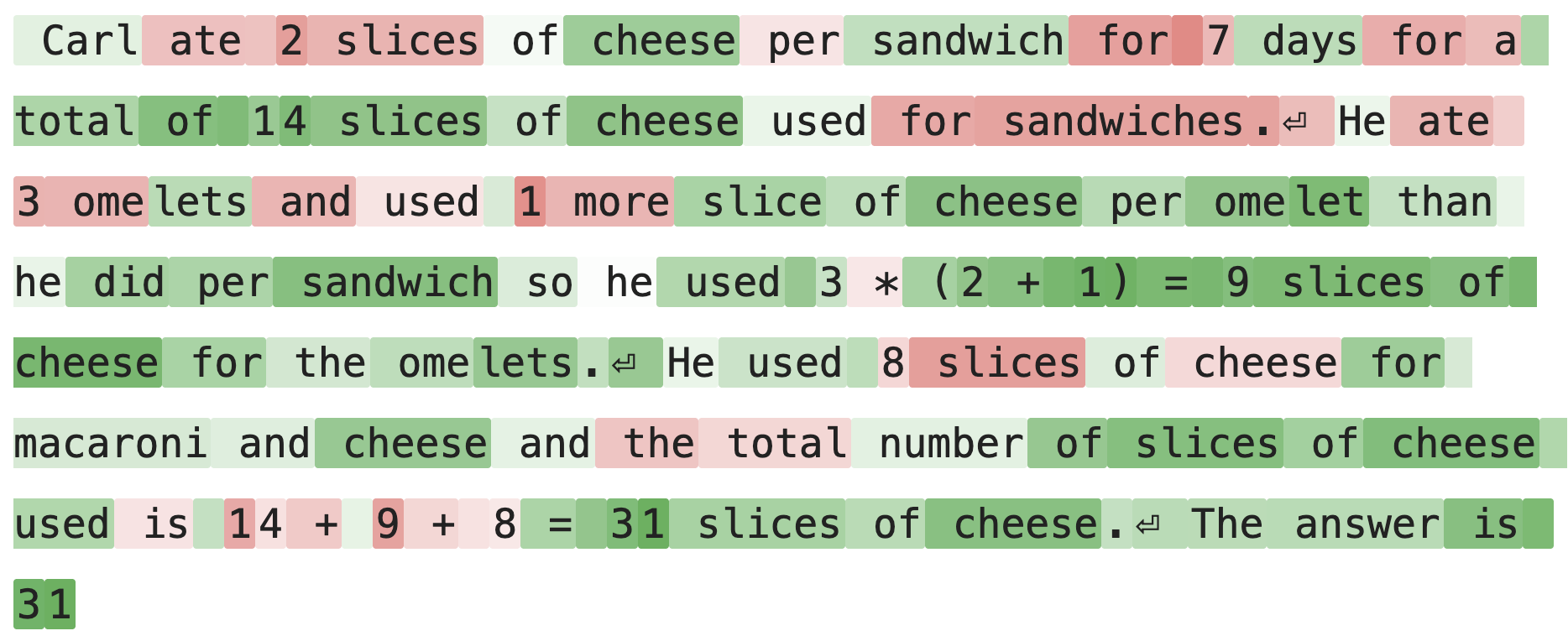}\\[-3pt]
\begin{minipage}{\columnwidth}
\tiny\textbf{Query 2}
\end{minipage}\\[1pt]
\includegraphics[width=\columnwidth]{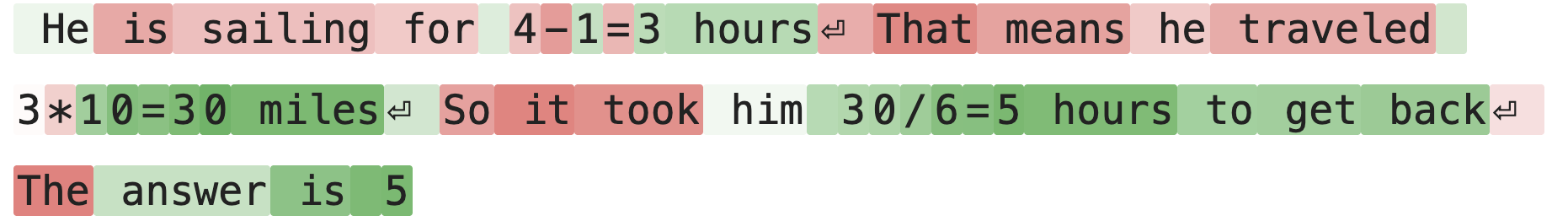}
\caption{\textbf{Token-level routing examples on GSM8K with Gemma3-4B.} Critical points routed to the expert are in red; base-generated tokens in green; soft-blend tokens in intermediate shades (approaching white as $m_t$ $\to$ 0.5). Additional examples in Appendix \ref{app:token}.}
\label{fig:viz_gsm_math}
\vspace{-5mm}
\end{figure}

\medskip\noindent\textbf{Token-level Analysis of Routing Behavior.}
To complement the quantitative invocation statistics, we visualize CPR's per-token routing decisions in Figure \ref{fig:viz_gsm_math}. 
One can observe that the expert is invoked predominantly on numeric tokens and arithmetic operators that drive multi-step calculation, while connective and explanatory tokens are left to the base. 
This token-level selectivity provides a mechanistic explanation for the invocation patterns shown in Figure \ref{fig:efficiency}: CPR preserves general capability not by uniformly damping expert influence, but by routing the expert precisely to tokens where domain knowledge is required.

\medskip\noindent\textbf{Generalization to External Experts.}
In our primary experiments, we evaluated CPR using domain experts fine-tuned via a controlled, vanilla SFT process. 
A natural question is whether CPR relies on this specific expert construction or if the framework generalizes to independently produced experts. 
To investigate this, we replace the SFT-trained expert with publicly available experts while keeping the base model, labeling procedure, router training, and all inference-time settings unchanged. 
Specifically, we consider two configurations on math task:
\begin{itemize}[leftmargin=*, itemsep=0pt, topsep=0pt]
    \item[$\circ$] \textit{OpenMath2-Llama3.1-8B} \citep{toshniwal2025openmathinstruct} paired with the Llama3.1-8B base model.
    \item[$\circ$] \textit{Qwen2.5-Math-1.5B} \citep{yang2024qwen2}, a smaller-scale math expert, paired with the Qwen2.5-1.5B base model.
\end{itemize} 
As shown in Table \ref{table:ext_expert}, the external experts exhibit the same pattern of catastrophic forgetting observed in our main experiments. 
Across both settings, the external experts replicate the forgetting pattern from our main experiments: general capability drops by 18.2\% (Llama3.1-8B) and 23.3\% (Qwen2.5-1.5B), with the latter even resulting in a net overall -7.8\% loss. 
In contrast, CPR recovers these drops to +0.7\% and -0.6\% respectively, while retaining most of the domain gains, yielding overall improvements of +17.7\% and +1.8\% over the respective base models. 
Notably, CPR makes the Qwen expert beneficial even when it underperforms the base on its own.

Overall, these results clearly demonstrate that CPR is not inherently tied to a specific SFT setting. 
As long as the expert outperforms the base model on a meaningful subset of tokens, the router can successfully identify these critical points and dispatch generation accordingly. 
This decoupling from the expert's training pipeline suggests that CPR can serve as a post-hoc remedy applicable to arbitrary third-party domain experts.

\section{Conclusion}

In this work, we propose \textbf{CPR} (\textbf{C}ritical-\textbf{P}oint \textbf{R}outing), a token-level routing framework that mitigates catastrophic forgetting in domain-specific SFT, by decoupling domain and general capabilities at the model level. 
A lightweight hierarchical router, trained on automatically labeled critical points, invokes the SFT expert only where the base model fails, while momentum smoothing and threshold-gated 3-way dispatch stabilize inference.
Across four model-domain settings, CPR consistently surpasses SFT in domain performance while fully recovering its general-capability drop.
In addition, CPR selectively invokes the expert, substantially reducing latency compared with always-on collaborative decoding.  
Token-level analyses further show that expert calls concentrate on domain-relevant tokens, providing a mechanistic view of how CPR preserves general capability while retaining domain specialization. 
CPR also generalizes to external experts without additional SFT. 
These results indicate that catastrophic forgetting is better addressed as an architectural problem than as a loss-level trade-off.

\section*{Limitations}
CPR effectively decouples domain and general capabilities, yet requires holding both the base model and the domain expert simultaneously, increasing GPU memory overhead relative to single-model approaches. 
In addition, although the expert is invoked on only \textasciitilde30\% of tokens overall, the base model must always run to produce the hidden states that feed the router, leaving residual per-step latency above single-model inference; complementary techniques such as prefix-only routing may further reduce this cost.
Beyond this, each expert invocation after skipped tokens incurs a key-value cache catch-up cost, yet CPR remains faster than ensemble and contrastive decoding since LLM decoding is memory-bound and weight loads are amortized across catch-up tokens within a single forward call.
CPR also assumes access to an expert that outperforms the base on a meaningful subset of tokens for constructing critical-point labels, so its effectiveness may degrade when this gap is narrow or when domain training data is scarce. 

\section*{Broader Impact and Ethical Implications}
CPR offers a practical way to deploy domain-specialized LLMs without sacrificing general capabilities, which is particularly valuable in high-stakes domains where brittleness in general reasoning can lead to harmful outputs. 
However, CPR inherits the biases and failure modes of both the base model and the domain expert model, and routing does not correct errors common to both. 
Since CPR can be applied post-hoc to arbitrary third-party experts, careful vetting of expert models is essential before deployment. 
The medical domain experiments in this work are intended solely for research evaluation on public benchmarks and do not constitute clinical validation. 
All datasets and models used are publicly available and were used in accordance with their respective licenses.

\section*{Acknowledgement}
Jaehyung Kim is affiliated with the Department of Artificial Intelligence at Yonsei University.
This research was supported in part by Institute for Information \& communications Technology Planning \& Evaluation (IITP) grant funded by the Korea government (MSIT) (No. RS-2020-II201361, Artificial Intelligence Graduate School Program (Yonsei University); No. RS-2026-25522672, Development of Unified Reasoning Technology Mimicking Human Cognition for Hierarchical Understanding and Unbounded Problem Solving).
Yunhun Nam and Jongheon Jeong acknowledge support from IITP grants funded by the Korea government (MSIT) (No.~RS-2019-II190079, Artificial Intelligence Graduate School Program (Korea University); No.~IITP-2026-RS-2025-02304828, Artificial Intelligence Star Fellowship Support Program to Nurture the Best Talents).

\bibliography{custom}

@STRING{iclr = {International Conference on Learning Representations (ICLR)}}

@STRING{neurips = {Advances in Neural Information Processing Systems (NeurIPS)}}

@STRING{tmlr = {Transactions on Machine Learning Research (TMLR)}}

@STRING{emnlp = {Conference on Empirical Methods in Natural Language Processing (EMNLP)}}

@STRING{acl = {Annual Meeting of the Association for Computational Linguistics (ACL)}}

@STRING{naacl = {Conference of the North American Chapter of the Association for Computational Linguistics: Human Language Technologies (NAACL-HLT)}}

@inproceedings{wei2022chain,
  title={Chain-of-thought prompting elicits reasoning in large language models},
  author={Wei, Jason and Wang, Xuezhi and Schuurmans, Dale and Bosma, Maarten and Xia, Fei and Chi, Ed and Le, Quoc V and Zhou, Denny and others},
  booktitle=neurips,
  volume={35},
  pages={24824--24837},
  year={2022}
}

@inproceedings{kojima2022large,
  title={Large language models are zero-shot reasoners},
  author={Kojima, Takeshi and Gu, Shixiang Shane and Reid, Machel and Matsuo, Yutaka and Iwasawa, Yusuke},
  booktitle=neurips,
  volume={35},
  pages={22199--22213},
  year={2022}
}

@inproceedings{toshniwal2025openmathinstruct,
  title={Openmathinstruct-2: Accelerating ai for math with massive open-source instruction data},
  author={Toshniwal, Shubham and Du, Wei and Moshkov, Ivan and Kisacanin, Branislav and Ayrapetyan, Alexan and Gitman, Igor},
  booktitle=iclr,
  volume={2025},
  pages={19243--19275},
  year={2025}
}

@article{han2023medalpaca,
  title={MedAlpaca--an open-source collection of medical conversational AI models and training data},
  author={Han, Tianyu and Adams, Lisa C and Papaioannou, Jens-Michalis and Grundmann, Paul and Oberhauser, Tom and Figueroa, Alexei and L{\"o}ser, Alexander and Truhn, Daniel and Bressem, Keno K},
  journal={arXiv preprint arXiv:2304.08247},
  year={2023}
}

@article{luo2025empirical,
  title={An empirical study of catastrophic forgetting in large language models during continual fine-tuning},
  author={Luo, Yun and Yang, Zhen and Meng, Fandong and Li, Yafu and Zhou, Jie and Zhang, Yue},
  journal={IEEE Transactions on Audio, Speech and Language Processing},
  year={2025},
  publisher={IEEE}
}

@inproceedings{kotha2024understanding,
  title={Understanding catastrophic forgetting in language models via implicit inference},
  author={Kotha, Suhas and Springer, Jacob and Raghunathan, Aditi},
  booktitle=iclr,
  volume={2024},
  pages={24110--24139},
  year={2024}
}

@inproceedings{liu2024more,
  title={More than catastrophic forgetting: Integrating general capabilities for domain-specific llms},
  author={Liu, Chengyuan and Kang, Yangyang and Wang, Shihang and Qing, Lizhi and Zhao, Fubang and Wu, Chao and Sun, Changlong and Kuang, Kun and Wu, Fei},
  booktitle=emnlp,
  pages={7531--7548},
  year={2024}
}

@inproceedings{wu2025generalization,
  title={On the generalization of sft: A reinforcement learning perspective with reward rectification},
  author={Wu, Yongliang and Zhou, Yizhou and Ziheng, Zhou and Peng, Yingzhe and Ye, Xinyu and Hu, Xinting and Zhu, Wenbo and Qi, Lu and Yang, Ming-Hsuan and Yang, Xu},
  booktitle=iclr,
  year={2026}
}

@article{diao2026entropy,
  title={Entropy-Adaptive Fine-Tuning: Resolving Confident Conflicts to Mitigate Forgetting},
  author={Diao, Muxi and Yang, Lele and Gong, Wuxuan and Zhang, Yutong and Yan, Zhonghao and Han, Yufei and Liang, Kongming and Xu, Weiran and Ma, Zhanyu},
  journal={arXiv preprint arXiv:2601.02151},
  year={2026}
}

@inproceedings{nam2026learning,
  title={Learning from the Undesirable: Robust Adaptation of Language Models without Forgetting},
  author={Nam, Yunhun and Kim, Jaehyung and Jeong, Jongheon},
  booktitle={Proceedings of the AAAI Conference on Artificial Intelligence},
  volume={40},
  number={38},
  pages={32537--32545},
  year={2026}
}

@inproceedings{lin2025sft,
  title={SFT Doesn't Always Hurt General Capabilities: Revisiting Domain-Specific Fine-Tuning in LLMs},
  author={Lin, Jiacheng and Wang, Zhongruo and Qian, Kun and Wang, Tian and Srinivasan, Arvind and Zeng, Hansi and Jiao, Ruochen and Zhou, Xie and Gesi, Jiri and Wang, Dakuo and others},
  booktitle=iclr,
  year={2026}
}

@inproceedings{ong2024routellm,
  title={Routellm: Learning to route llms with preference data},
  author={Ong, Isaac and Almahairi, Amjad and Wu, Vincent and Chiang, Wei-Lin and Wu, Tianhao and Gonzalez, Joseph E and Kadous, M Waleed and Stoica, Ion},
  booktitle=iclr,
  year={2025}
}

@inproceedings{chen2023frugalgpt,
  title={Frugalgpt: How to use large language models while reducing cost and improving performance},
  author={Chen, Lingjiao and Zaharia, Matei and Zou, James},
  booktitle=tmlr,
  year={2024}
}

@inproceedings{ding2024hybrid,
  title={Hybrid llm: Cost-efficient and quality-aware query routing},
  author={Ding, Dujian and Mallick, Ankur and Wang, Chi and Sim, Robert and Mukherjee, Subhabrata and R{\"u}hle, Victor and Lakshmanan, Laks and Awadallah, Ahmed H},
  booktitle=iclr,
  volume={2024},
  pages={41348--41366},
  year={2024}
}

@inproceedings{feng2025don,
  title={Don't Throw Away Your Pretrained Model},
  author={Feng, Shangbin and Yu, Wenhao and Wang, Yike and Zhang, Hongming and Tsvetkov, Yulia and Yu, Dong},
  booktitle=iclr,
  year={2026}
}

@article{xiong2026token,
  title={Token-Level LLM Collaboration via FusionRoute},
  author={Xiong, Nuoya and Zhou, Yuhang and Zeng, Hanqing and Chen, Zhaorun and Huang, Furong and Bi, Shuchao and Zhang, Lizhu and Zhao, Zhuokai},
  journal={arXiv preprint arXiv:2601.05106},
  year={2026}
}

@inproceedings{shen2024learning,
  title={Learning to decode collaboratively with multiple language models},
  author={Shen, Zejiang and Lang, Hunter and Wang, Bailin and Kim, Yoon and Sontag, David},
  booktitle=acl,
  pages={12974--12990},
  year={2024}
}

@inproceedings{li2023contrastive,
  title={Contrastive decoding: Open-ended text generation as optimization},
  author={Li, Xiang Lisa and Holtzman, Ari and Fried, Daniel and Liang, Percy and Eisner, Jason and Hashimoto, Tatsunori B and Zettlemoyer, Luke and Lewis, Mike},
  booktitle=acl,
  pages={12286--12312},
  year={2023}
}

@article{Kamath2025Gemma3T,
  title={Gemma 3 Technical Report},
  author={Gemma Team and others},
  journal={arXiv preprint arXiv:2503.19786},
  year={2025}
}

@article{grattafiori2024llama,
  title={The llama 3 herd of models},
  author={Grattafiori, Aaron and Dubey, Abhimanyu and Jauhri, Abhinav and Pandey, Abhinav and Kadian, Abhishek and Al-Dahle, Ahmad and Letman, Aiesha and Mathur, Akhil and Schelten, Alan and Vaughan, Alex and others},
  journal={arXiv preprint arXiv:2407.21783},
  year={2024}
}

@article{cobbe2021training,
  title={Training verifiers to solve math word problems},
  author={Cobbe, Karl and Kosaraju, Vineet and Bavarian, Mohammad and Chen, Mark and Jun, Heewoo and Kaiser, Lukasz and Plappert, Matthias and Tworek, Jerry and Hilton, Jacob and Nakano, Reiichiro and others},
  journal={arXiv preprint arXiv:2110.14168},
  year={2021}
}

@inproceedings{jin2019pubmedqa,
  title={Pubmedqa: A dataset for biomedical research question answering},
  author={Jin, Qiao and Dhingra, Bhuwan and Liu, Zhengping and Cohen, William and Lu, Xinghua},
  booktitle=emnlp,
  pages={2567--2577},
  year={2019}
}

@article{yang2024qwen2,
  title={Qwen2.5-math technical report: Toward mathematical expert model via self-improvement},
  author={Yang, An and Zhang, Beichen and Hui, Binyuan and Gao, Bofei and Yu, Bowen and Li, Chengpeng and Liu, Dayiheng and Tu, Jianhong and Zhou, Jingren and Lin, Junyang and others},
  journal={arXiv preprint arXiv:2409.12122},
  year={2024}
}

@inproceedings{lobo2025impact,
  title={On the impact of fine-tuning on chain-of-thought reasoning},
  author={Lobo, Elita and Agarwal, Chirag and Lakkaraju, Himabindu},
  booktitle=naacl,
  pages={11679--11698},
  year={2025}
}

@inproceedings{hendrycks2020measuring,
  title={Measuring massive multitask language understanding},
  author={Hendrycks, Dan and Burns, Collin and Basart, Steven and Zou, Andy and Mazeika, Mantas and Song, Dawn and Steinhardt, Jacob},
  booktitle=iclr,
  year={2021}
}

@inproceedings{talmor2019commonsenseqa,
  title={Commonsenseqa: A question answering challenge targeting commonsense knowledge},
  author={Talmor, Alon and Herzig, Jonathan and Lourie, Nicholas and Berant, Jonathan},
  booktitle=naacl,
  pages={4149--4158},
  year={2019}
}

@article{clark2018think,
  title={Think you have solved question answering? try arc, the ai2 reasoning challenge},
  author={Clark, Peter and Cowhey, Isaac and Etzioni, Oren and Khot, Tushar and Sabharwal, Ashish and Schoenick, Carissa and Tafjord, Oyvind},
  journal={arXiv preprint arXiv:1803.05457},
  year={2018}
}

@inproceedings{miao2020diverse,
  title={A diverse corpus for evaluating and developing English math word problem solvers},
  author={Miao, Shen-Yun and Liang, Chao-Chun and Su, Keh-Yih},
  booktitle=acl,
  pages={975--984},
  year={2020}
}

@inproceedings{patel2021nlp,
  title={Are NLP models really able to solve simple math word problems?},
  author={Patel, Arkil and Bhattamishra, Satwik and Goyal, Navin},
  booktitle=naacl,
  pages={2080--2094},
  year={2021}
}

@article{jin2021disease,
  title={What disease does this patient have? a large-scale open domain question answering dataset from medical exams},
  author={Jin, Di and Pan, Eileen and Oufattole, Nassim and Weng, Wei-Hung and Fang, Hanyi and Szolovits, Peter},
  journal={Applied Sciences},
  volume={11},
  number={14},
  pages={6421},
  year={2021},
  publisher={MDPI}
}

@inproceedings{arias2025automatic,
  title={Automatic evaluation of healthcare LLMs beyond question-answering},
  author={Arias-Duart, Anna and Martin-Torres, Pablo Agustin and Hinjos, Daniel and Bernabeu-Perez, Pablo and Ganzabal, Lucia Urcelay and Mallo, Marta Gonzalez and Gururajan, Ashwin Kumar and Lopez-Cuena, Enrique and Alvarez-Napagao, Sergio and Garcia-Gasulla, Dario},
  booktitle=naacl,
  pages={108--130},
  year={2025}
}

@inproceedings{dou2024loramoe,
  title={LoRAMoE: Alleviating world knowledge forgetting in large language models via MoE-style plugin},
  author={Dou, Shihan and Zhou, Enyu and Liu, Yan and Gao, Songyang and Shen, Wei and Xiong, Limao and Zhou, Yuhao and Wang, Xiao and Xi, Zhiheng and Fan, Xiaoran and others},
  booktitle=acl,
  pages={1932--1945},
  year={2024}
}

@inproceedings{wu2024mixture,
  title={Mixture of lora experts},
  author={Wu, Xun and Huang, Shaohan and Wei, Furu},
  booktitle=iclr,
  year={2024}
}

@inproceedings{chen2021finqa,
  title={Finqa: A dataset of numerical reasoning over financial data},
  author={Chen, Zhiyu and Chen, Wenhu and Smiley, Charese and Shah, Sameena and Borova, Iana and Langdon, Dylan and Moussa, Reema and Beane, Matt and Huang, Ting-Hao and Routledge, Bryan R and others},
  booktitle=emnlp,
  pages={3697--3711},
  year={2021}
}

@article{dubois2024length,
  title={Length-controlled alpacaeval: A simple way to debias automatic evaluators},
  author={Dubois, Yann and Galambosi, Bal{\'a}zs and Liang, Percy and Hashimoto, Tatsunori B},
  journal={arXiv preprint arXiv:2404.04475},
  year={2024}
}

@misc{alpaca,
  author = {Rohan Taori and Ishaan Gulrajani and Tianyi Zhang and Yann Dubois and Xuechen Li and Carlos Guestrin and Percy Liang and Tatsunori B. Hashimoto },
  title = {Stanford Alpaca: An Instruction-following LLaMA model},
  year = {2023},
  publisher = {GitHub},
  journal = {GitHub repository},
  howpublished = {\url{https://github.com/tatsu-lab/stanford_alpaca}},
}
\clearpage
\appendix
\section{More Details of Experimental Setups}
\label{app:setups}
This section provides additional details about the experiments in Section~\ref{sec4}, organized as follows: datasets (Appendix~\ref{app:datasets}), baselines (Appendix~\ref{app:baselines}), implementation details (Appendix~\ref{app:implementation}), and prompt templates (Appendix \ref{app:prompts}).

\subsection{Datasets}
\label{app:datasets}

\paragraph{Training Data.}
We construct domain experts via vanilla SFT on two domains, math and medical.
\begin{itemize}
    \item \textbf{Math Domain.} We use the official training split of GSM8K \citep{cobbe2021training}, which consists of approximately 8K grade-school math word problems with step-by-step rationales. Training instances are formatted as 0-shot chain-of-thought (CoT), with the rationale and the final boxed answer as the target.
    \item \textbf{Medical Domain.} We sample from the artificial split of PubMedQA \citep{jin2019pubmedqa}, a biomedical research question-answering dataset whose answers are categorized as \texttt{yes}, \texttt{no}, or \texttt{maybe}. To prevent label-distribution bias and to match the math training size, we sample 4{,}000 \texttt{yes} and 4{,}000 \texttt{no} examples (8K total). Training instances are formatted as 2-shot CoT with the answer options randomly shuffled to prevent positional bias, and the 2-shot demonstrations are drawn from MMLU \citep{hendrycks2020measuring} so that the medical expert learns to use a CoT format that transfers naturally to multiple-choice evaluation.
\end{itemize}

\paragraph{Evaluation Benchmarks.}
We evaluate on the full test set of every benchmark used in this work, without any subsampling. The benchmarks are grouped into three categories.

\begin{itemize}
    \item \textbf{GSM8K} \citep{cobbe2021training}: grade-school math word problems requiring multi-step arithmetic reasoning; in-domain for the math expert.
    \item \textbf{ASDiv} \citep{miao2020diverse}: diverse English math word problems covering a broader range of problem types than GSM8K.
    \item \textbf{SVAMP} \citep{patel2021nlp}: math word problems constructed by applying simple variations to existing problems to test robustness.
\end{itemize}

\begin{itemize}
    \item \textbf{PubMedQA} \citep{jin2019pubmedqa}: biomedical yes/no/maybe question answering on research abstracts; in-domain for the medical expert.
    \item \textbf{MedQA} \citep{jin2021disease}: multiple-choice questions from professional medical board examinations.
    \item \textbf{CareQA} \citep{arias2025automatic}: a healthcare LLM evaluation suite spanning multiple clinical and biomedical topics.
\end{itemize}

\begin{itemize}
    \item \textbf{MMLU} \citep{hendrycks2020measuring}: a broad multitask evaluation suite spanning 57 subjects from humanities to STEM.
    \item \textbf{CommonsenseQA} \citep{talmor2019commonsenseqa}: multiple-choice questions targeting commonsense reasoning.
    \item \textbf{ARC-C} \citep{clark2018think}: the Challenge split of the AI2 Reasoning Challenge, consisting of grade-school science questions filtered to be difficult for retrieval-based methods.
\end{itemize}

\paragraph{Evaluation Protocol.}
Following the same protocol as training, math benchmarks (GSM8K, ASDiv, SVAMP) are evaluated with 0-shot CoT prompting, while medical (PubMedQA, MedQA, CareQA) and general (MMLU, CommonsenseQA, ARC-C) benchmarks are evaluated with 2-shot CoT prompting. The 2-shot CoT demonstrations are drawn from MMLU and held fixed across all evaluated methods within a setting, ensuring a controlled comparison. For multiple-choice medical benchmarks, the answer options follow the natural ordering of the benchmark; the random shuffling of options is applied only during medical SFT training to prevent positional bias in the expert. The exact prompt templates used for each benchmark are provided, in Appendix \ref{app:prompts}.

\subsection{Baselines}
\label{app:baselines}
This section provides additional details on the baselines compared in our experiments (Section~\ref{sec4}). All baselines share the same base model $f_B$ and the same vanilla SFT expert $f_E$ for a controlled comparison, and are organized into two groups: \emph{single-model baselines} that integrate domain and general capabilities into one weight set, and \emph{multi-model collaboration baselines} that route or combine the base and the expert.

\paragraph{Single-model Baselines.}
\begin{itemize}
    \item \textbf{Base} refers to the original pre-trained LLM without any domain adaptation (Gemma3-4B or Llama3.1-8B). It serves as the reference for general capability and the upper bound of forgetting-free behavior.

    \item \textbf{SFT (Expert)} is the domain expert obtained by vanilla supervised fine-tuning of the base model on the domain training set (GSM8K for math; PubMedQA artificial split for medical). It serves as the reference for domain capability and the lower bound for general capability under standard SFT.

    \item \textbf{DFT} \citep{wu2025generalization} rescales the per-token cross-entropy loss by the model's own predicted probability of the target token, with the goal of neutralizing an implicit inverse-probability reward that standard SFT gradients encode. This single-line modification of the SFT loss is shown to improve generalization relative to vanilla SFT. We include DFT as a representative regularization-based method that modifies the SFT loss at the token level. It shares all training hyperparameters with vanilla SFT.

    \item \textbf{EAFT} \citep{diao2026entropy} addresses catastrophic forgetting by identifying \emph{Confident Conflict} tokens, where the model assigns low probability and low entropy to the ground-truth token (i.e., it is confident in its own divergent prediction). EAFT uses token-level entropy as a soft gate: tokens with low entropy are down-weighted to suppress destructive gradient updates, while tokens with high entropy retain the standard SFT signal to enable genuine learning. We include EAFT as a state-of-the-art entropy-guided regularization baseline. Training hyperparameters are identical to vanilla SFT.

    \item \textbf{LfU} \citep{nam2026learning} regularizes the SFT process to favor solutions that are resilient to \emph{undesirable} updates, simulated by a one-step gradient-ascent direction on an auxiliary model. It enforces consistency between the internal representations of the original model and those of the undesirably-updated model, encouraging robust adaptation without forgetting. The original paper provides two auxiliary-model variants: a LoRA-based variant and a representation-steering variant. We adopt the LoRA-based variant, which generally yields stronger performance in the original paper, and follow the default LoRA configuration and training hyperparameters reported therein.
\end{itemize}

\paragraph{Multi-model Collaboration Baselines.}

\begin{itemize}

    \item \textbf{RouteLLM} \citep{ong2024routellm} dispatches an entire query to either a strong or a weak LLM based on a router trained on human preference data. Among the four router variants proposed in the original paper, we adopt the Causal LLM classifier, which is reported as the strongest approach on GSM8K when trained on preference data augmented with an LLM judge. We use the officially released checkpoint \texttt{routellm/causal\_\allowbreak llm\_\allowbreak gpt4\_\allowbreak augmented} on Hugging Face, trained on Chatbot Arena preference data augmented with GPT-4 judgments, which the original paper reports to outperform the Arena-only variant. The router is used without any re-training to ensure faithful reproduction, with $f_B$ and $f_E$ substituted as the weak and strong models, respectively. RouteLLM represents the \emph{query-level} routing granularity.

    \item \textbf{Switch Generation} \citep{feng2025don} alternates between multiple checkpoints (e.g., pre-trained, fine-tuned, aligned) at the patch level during generation, recovering skills that may have been lost during alignment or domain adaptation. We use the officially released \texttt{bunsenfeng/PFA\_\allowbreak switcher\_\allowbreak 1} switcher that learned from diverse tasks, contexts, and model collaboration pattern, applied without re-training. Switch Generation represents the \emph{patch-level} routing granularity, situated between query-level dispatch and our token-level routing.

    \item \textbf{Ensemble} computes the next-token distribution as a uniform average of the base and expert distributions, $p_t = \tfrac{1}{2}(p_b + p_e)$, at every decoding step. It serves as the simplest collaborative-decoding baseline and provides a reference point for fixed-weight mixture decoding.

    \item \textbf{Contrastive Decoding} \citep{li2023contrastive} computes the next-token distribution by subtracting the log-probability of an \emph{amateur} LM from that of an \emph{expert} LM, subject to a plausibility constraint: $\log p_t \;\propto\; \log p_{\text{exp}} - \log p_{\text{ama}}$. Following the assumption that the SFT model is stronger on the target domain than the base, we fix the SFT expert $f_E$ as the expert and the base $f_B$ as the amateur in all settings and use the default threshold of $\alpha = 0.1$ for the plausibility constraint.
    Both Ensemble and Contrastive Decoding invoke $f_B$ and $f_E$ at every decoding step, doubling per-token compute, in contrast to CPR's selective expert invocation.
    
\end{itemize}

\subsection{Implementation Details}
\label{app:implementation}
This section provides the detailed information needed to reproduce our main experiments.

\paragraph{Compute.}
All experiments are conducted on NVIDIA RTX A6000 GPUs. Each model fits on a single A6000, and we parallelize over multiple A6000s to run different models concurrently.

\paragraph{SFT and Full Fine-tuning Baselines.}
We re-implement DFT and EAFT under matched training settings with the vanilla SFT expert to ensure a controlled comparison. All three share identical hyperparameters across both domains: maximum sequence length 1024, per-device batch size 4, gradient accumulation steps 4 (effective batch size 16), learning rate $2\mathrm{e}{-5}$, constant learning-rate schedule, and 6 training epochs.

\paragraph{LoRA-based Baseline.}
For LfU, we re-implement it following the default LoRA configuration and training hyperparameters of the original paper \citep{nam2026learning}.

\paragraph{Router Training (CPR).}
We freeze both $f_B$ and $f_E$ and train only the hierarchical router parameters $\phi$. The router's hidden dimension is $d_h = 256$ for both the macro encoder and the micro router, both implemented as 2-layer ReLU MLPs (Section~\ref{sec3}). Training uses maximum sequence length 1024, batch size 16, learning rate $1\mathrm{e}{-4}$, warmup ratio 0.1, and 6 epochs, applied uniformly across both domains. For prompts that include few-shot demonstrations, the macro encoder pooling is restricted to the question span so that the macro summary reflects the domain of the actual query rather than the fixed few-shot demonstrations, which are shared across queries within a benchmark and would otherwise dilute the domain signal.

\paragraph{Critical-point Labeling.}
Critical-point labels are constructed via teacher-forcing comparison between $f_B$ and $f_E$ on the \emph{same training data} used for vanilla SFT (Appendix~\ref{app:datasets}), i.e., GSM8K for math and the balanced PubMedQA sample for medical. At each generation step, we compare the greedy predictions of both models against the ground-truth token and assign labels $z_t \in \{0, 1, \emptyset\}$ as defined in Eq.\ref{eq:labeling}. No additional data beyond the SFT training set is used to construct router supervision.

\paragraph{Inference Settings.}
All compared methods (CPR and all baselines) share the same inference configuration to control for the generation environment: greedy decoding with temperature $0$, with a sufficiently large maximum generation length to avoid truncation. For routing-based baselines (RouteLLM, Switch Generation), we use the officially released router checkpoints without any re-training. For CPR, we use the default values $\alpha = 0.5$, $\tau_{\text{low}} = 0.35$, and $\tau_{\text{high}} = 0.65$ unless otherwise specified.

\subsection{Prompt Templates}
\label{app:prompts}
We provide the exact prompt templates used for each benchmark in 
Listings \ref{lst:prompt-gsm8k}-\ref{lst:prompt-arcc}. Math benchmarks 
(GSM8K, ASDiv, SVAMP) use 0-shot CoT prompting, while medical (PubMedQA, 
MedQA, CareQA) and general (MMLU, CommonsenseQA, ARC-C) benchmarks use 
2-shot CoT with demonstrations drawn from MMLU and held fixed across 
all methods. For 2-shot demonstrations, we select examples from 
MMLU categories relevant to each benchmark domain: medical-related 
categories (e.g., clinical knowledge, medical genetics, anatomy) for 
medical benchmarks, and a broad mix of 
categories for general benchmarks. Within each domain-relevant category 
pool, demonstrations are sampled randomly and held fixed across all 
compared methods. The same templates are applied consistently throughout 
all stages of our pipeline including SFT training, router-training critical-point labeling, and evaluation, ensuring no prompt mismatch between training 
and evaluation.

\begin{itemize}
    \item \textbf{Math Benchmarks (0-shot CoT):} 
    GSM8K (Listing~\ref{lst:prompt-gsm8k}), 
    ASDiv (Listing~\ref{lst:prompt-asdiv}), 
    SVAMP (Listing~\ref{lst:prompt-svamp}).
    \item \textbf{Medical Benchmarks (2-shot CoT):} 
    PubMedQA (Listing~\ref{lst:prompt-pubmedqa}), 
    MedQA (Listing~\ref{lst:prompt-medqa}), 
    CareQA (Listing~\ref{lst:prompt-careqa}).
    \item \textbf{General Benchmarks (2-shot CoT):} 
    MMLU (Listing~\ref{lst:prompt-mmlu}), 
    CommonsenseQA (Listing~\ref{lst:prompt-csqa}), 
    ARC-C (Listing~\ref{lst:prompt-arcc}).
\end{itemize}

\begin{table*}[t]
\renewcommand{\arraystretch}{1.0}
\centering
\scriptsize
\setlength{\tabcolsep}{4pt}
\resizebox{\textwidth}{!}{%
\begin{tabular}{lccccccccc}
\toprule

\multicolumn{1}{c}{} 
& \multicolumn{4}{c}{\textbf{Math Domain}} 
& \multicolumn{4}{c}{\textbf{General Domain}} 
& \multirow{2}{*}{\textbf{Overall}} \\

\cmidrule(lr){2-5} \cmidrule(lr){6-9}

\textbf{Method}
& \textbf{GSM8K\textsuperscript{\(\star\)}}
& \textbf{ASDiv}
& \textbf{SVAMP}
& \textbf{Average}
& \textbf{MMLU}
& \textbf{CSQA}
& \textbf{ARC-C}
& \textbf{Average}
& \textbf{Average} \\

\midrule

\multicolumn{10}{c}{\textit{Gemma3-4B}} \\
\midrule

Base & 29.11 & 44.51 & 48.67 & 40.76 & 54.16 & 59.13 & 64.42 & 59.24 & 50.00 \\

Switch Generation-ours & 48.07 & 57.96 & 49.67 & 51.90 & 50.98 & 49.39 & 57.47 & 52.61 & 52.23 \\

Switch Generation-official & 46.93 & 57.43 & 49.67 & 51.34 & 50.90 & 52.90 & 57.25 & 53.68 & 52.51 \\

RouteLLM-ours & 51.33 & 57.61 & 50.00 & 52.98 & 47.97 & 48.81 & 56.31 & 51.03 & 52.01 \\

RouteLLM-official & 49.63 & 57.87 & 50.67 & 52.72 & 49.91 & 52.99 & 56.01 & 52.97 & 52.85 \\

CPR \textbf{(Ours)} & 49.58 & 62.60 & 56.00 & 56.06 & 55.89 & 61.10 & 67.24 & 61.41 & 58.74 \\

\bottomrule
\end{tabular}%
}
\caption{\textbf{Re-trained routing-based baselines.} Comparison between officially released router checkpoints and variants re-trained on our SFT data for RouteLLM and Switch Generation, evaluated on math domain for Gemma3-4B.} 
\label{table:own_router}
\vspace{-2mm}
\end{table*}

\begin{table}[t]
\centering
\small
\setlength{\tabcolsep}{6pt}
\begin{tabular}{llcc}
\toprule
\textbf{Domain} & \textbf{Benchmark} & \textbf{Gemma3-4B} & \textbf{Llama3.1-8B} \\
\midrule
\multirow{8}{*}{Math} 
  & GSM8K    & 34.8  & 37.9  \\
  & ASDiv    & 35.4  & 41.6  \\
  & SVAMP    & 31.8  & 40.2  \\
  & Average  & 34.00 & 39.90 \\
\cmidrule(lr){2-4}
  & MMLU     & 24.4  & 20.6  \\
  & CSQA     & 13.1  & 32.3  \\
  & ARC-C    & 27.3  & 35.5  \\
  & Average  & 21.60 & 29.47 \\
\midrule
\multirow{8}{*}{Medical} 
  & PubMedQA & 14.8  & 41.6  \\
  & MedQA    & 42.0  & 55.0  \\
  & CareQA   & 35.9  & 41.3  \\
  & Average  & 30.90 & 45.96 \\
\cmidrule(lr){2-4}
  & MMLU     & 22.6  & 27.7  \\
  & CSQA     & 12.8  & 43.9  \\
  & ARC-C    & 24.8  & 41.0  \\
  & Average  & 20.07 & 37.53 \\
\bottomrule
\end{tabular}
\caption{\textbf{Full benchmark numbers of expert invocation rate (\%).} 
Full benchmark numbers of Figure \ref{fig:efficiency}.}
\label{table:efficiency_full}
\end{table}
\begin{table*}[t]
\renewcommand{\arraystretch}{1.0}
\centering
\scriptsize
\setlength{\tabcolsep}{4pt}
\resizebox{\textwidth}{!}{%
\begin{tabular}{lccccccccc}
\toprule
\multicolumn{1}{c}{} 
& \multicolumn{4}{c}{\textbf{Math Domain}} 
& \multicolumn{4}{c}{\textbf{General Domain}} 
& \multirow{2}{*}{\textbf{Overall}} \\
\cmidrule(lr){2-5} \cmidrule(lr){6-9}
\textbf{Method}
& \textbf{GSM8K\textsuperscript{\(\star\)}}
& \textbf{ASDiv}
& \textbf{SVAMP}
& \textbf{Average}
& \textbf{MMLU}
& \textbf{CSQA}
& \textbf{ARC-C}
& \textbf{Average}
& \textbf{Average} \\
\midrule
\multicolumn{10}{c}{\textit{Gemma3-4B}} \\
\midrule
Base           
& 29.11 & 44.51 & 48.67 
& 40.76 
& 54.16 & 59.13 & 64.42 
& 59.24 
& 50.00 \\
w/o Macro
& 48.82 & 62.56 & 55.00 & 55.46 {\color{blue}(+14.70)}
& 54.90 & 61.83 & 62.06 & 59.60 {\color{gray}(+0.36)}
& 57.53 {\color{blue}(+7.53)} \\
w/o Micro
& 37.53 & 55.49 & 45.67 
& 46.23 {\color{blue}(+5.47)}
& 53.22 & 61.67 & 67.49 
& 60.79 {\color{blue}(+1.55)}
& 53.51 {\color{blue}(+3.51)}\\
Hierarchical \textbf{(Default)}
& 49.58 & 62.60 & 56.00 & \textbf{56.06} {\color{blue}(+15.30)}
& 55.89 & 61.10 & 67.24 & \textbf{61.41} {\color{blue}(+2.17)}
& \textbf{58.74} {\color{blue}(+8.74)} \\
\midrule
\multicolumn{10}{c}{\textit{Llama3.1-8B}} \\
\midrule
Base           
& 34.72 & 35.84 & 44.33 
& 38.30 
& 57.23 & 61.18 & 71.67 
& \textbf{63.36}
& 50.83 \\
w/o Macro
& 51.55 & 62.52 & 64.67 & 59.58 {\color{blue}(+21.28)}
& 58.74 & 64.62 & 61.86 & 61.74 {\color{red}(-1.62)}
& 60.66 {\color{blue}(+9.83)} \\
w/o Micro
& 43.52 & 58.87 & 56.66 
& 53.02 {\color{blue}(+14.72)}
& 55.38 & 66.26 & 63.65 
& 61.76 {\color{red}(-1.60)}
& 57.39 {\color{blue}(+6.56)}\\
Hierarchical \textbf{(Default)}
& 51.80 & 62.70 & 66.67 & \textbf{60.39} {\color{blue}(+22.09)}
& 58.97 & 64.26 & 65.25 & 62.83 {\color{gray}(-0.53)}
& \textbf{61.61} {\color{blue}(+10.78)} \\
\bottomrule
\end{tabular}%
}
\caption{\textbf{Ablation study on router architecture.} Effect of removing each component (macro encoder, micro router) from the hierarchical router, evaluated on math domain for both backbones.}
\label{table:ablation_router_math}
\vspace{-2mm}
\end{table*}
\begin{table*}[t]
\centering
\scriptsize
\setlength{\tabcolsep}{4pt}
\resizebox{\textwidth}{!}{%
\begin{tabular}{lccccccccc}
\toprule
\multicolumn{1}{c}{} 
& \multicolumn{4}{c}{\textbf{Medical Domain}} 
& \multicolumn{4}{c}{\textbf{General Domain}} 
& \multirow{2}{*}{\textbf{Overall}} \\
\cmidrule(lr){2-5} \cmidrule(lr){6-9}
\textbf{Method}
& \textbf{PubMedQA\textsuperscript{\(\star\)}}
& \textbf{MedQA}
& \textbf{CareQA}
& \textbf{Average}
& \textbf{MMLU}
& \textbf{CSQA}
& \textbf{ARC-C}
& \textbf{Average}
& \textbf{Average} \\
\midrule
\multicolumn{10}{c}{\textit{Gemma3-4B}} \\
\midrule
Base           
& 54.30 & 27.97 & 38.13 & 40.13 
& 54.16 & 59.13 & 64.42 & 59.24 
& 49.69 \\
w/o Macro
& 66.30 & 47.05 & 44.89 & 52.75 {\color{blue}(+12.62)}
& 55.01 & 60.85 & 69.27 & 61.71 {\color{blue}(+2.47)}
& 57.23 {\color{blue}(+7.54)} \\
w/o Micro
& 70.80 & 46.19 & 44.98 
& \textbf{53.99} {\color{blue}(+13.86)}
& 55.05 & 60.69 & 69.39 
& 61.71 {\color{blue}(+2.47)}
& \textbf{57.85} {\color{blue}(+8.16)}\\
Hierarchical \textbf{(Default)}
& 66.80 
& 48.20 
& 45.19 
& 53.40 {\color{blue}(+13.27)} 
& 54.82 
& 61.26 
& 69.67 
& \textbf{61.92} {\color{blue}(+2.68)} 
& 57.66 {\color{blue}(+7.97)} \\
\midrule
\multicolumn{10}{c}{\textit{Llama3.1-8B}} \\
\midrule
Base           
& 68.00 & 43.99 & 50.84 & 54.28 
& 57.23 & 61.18 & 71.67 & 63.36
& 58.82 \\
w/o Macro
& 75.90 & 53.10 & 56.36 & \textbf{61.79} {\color{blue}(+7.51)}
& 58.21 & 68.02 & 63.87 & 63.37 {\color{gray}(+0.01)}
& 62.58 {\color{blue}(+3.76)} \\
w/o Micro
& 76.20 & 52.39 & 55.29 
& 61.29 {\color{blue}(+7.01)}
& 57.61 & 69.51 & 61.08 
& 62.73 {\color{gray}(-0.63)}
& 62.01 {\color{blue}(+3.19)}\\
Hierarchical \textbf{(Default)}
& 74.60 
& 53.86 
& 56.62 
& 61.69 {\color{blue}(+7.41)}
& 58.89 
& 68.93 
& 65.27
& \textbf{64.36} {\color{blue}(+1.00)}
& \textbf{63.03} {\color{blue}(+4.21)} \\
\bottomrule
\end{tabular}%
}
\caption{\textbf{Ablation study on router architecture.} Same setting as Table \ref{table:ablation_router_math} on the medical domain.}
\label{table:ablation_router_medical}
\vspace{-2mm}
\end{table*}

\section{Additional Quantitative Results}
\label{app:quantitative}
This section provides additional quantitative results that complement the main experiments in Section \ref{sec4}. 
We report a comparison between officially released and re-trained routing baselines (Appendix \ref{app:retrained-routers}), full benchmark-level expert invocation rates in Figure \ref{fig:efficiency} (Appendix \ref{app:efficiency}), full per-benchmark numbers for the ablation studies summarized in Table \ref{table:ablation_mini} (Appendix \ref{app:ablation}), comparison with MoE-style routing baselines (Appendix~\ref{app:moe}), evaluation on the finance domain (Appendix~\ref{app:finance}) and open-ended instruction following setting (Appendix~\ref{app:alpacaeval}), and a lightweight LoRA configuration (Appendix~\ref{app:lora}).

\subsection{Re-trained Routing-based Baselines}
\label{app:retrained-routers}

The routing-based baselines in our main experiments (RouteLLM, Switch Generation) use officially released checkpoints. To examine whether re-training their routers on our SFT data could improve performance, we re-trained both following the settings specified in each paper, except that the rollout for Switch Generation was set to 8 due to computational constraints. As shown in Table~\ref{table:own_router}, the re-trained variants underperform the official checkpoints on both baselines, likely because our \textasciitilde8K SFT-scale data is insufficient to learn an effective query-level or patch-level router. We therefore adopt the official checkpoints in the main results (Tables~\ref{table:main_math}-\ref{table:main_medical}).

\subsection{Additional Efficiency Results}
\label{app:efficiency}
Table \ref{table:efficiency_full} provides the per-benchmark breakdown of the expert invocation rates aggregated in Figure \ref{fig:efficiency}. The rate remains around one-third of tokens across all benchmarks, and a clear in-domain vs. out-of-domain gap is observed in every setting, confirming that the router concentrates expert calls where domain knowledge is genuinely required.

\subsection{Additional Ablation Results}
\label{app:ablation}
Table~\ref{table:ablation_mini} in the main text reports averages aggregated across two backbones in the math setting. Tables~\ref{table:ablation_router_math}-\ref{table:ablation_threshold} provide the full per-benchmark numbers, separately for math and medical domains and for both backbones. The conclusions in Section~\ref{sec4.3.ablation} hold across all benchmarks: (i) the hierarchical router design, combining a query-level macro encoder and a token-level micro router, outperforms either component alone, (ii) $\alpha = 0.5$ outperforms both $\alpha = 0.0$ and $\alpha = 0.9$, and (iii) the 3-way dispatch with $(\tau_{low}, \tau_{high}) = (0.35, 0.65)$ outperforms both hard switching $(0.5, 0.5)$ and fully soft blending $(0.0, 1.0)$.

\begin{table*}[t]
\renewcommand{\arraystretch}{1.0}
\centering
\scriptsize
\setlength{\tabcolsep}{4pt}
\resizebox{\textwidth}{!}{%
\begin{tabular}{lccccccccc}
\toprule

\multicolumn{1}{c}{} 
& \multicolumn{4}{c}{\textbf{Math Domain}} 
& \multicolumn{4}{c}{\textbf{General Domain}} 
& \multirow{2}{*}{\textbf{Overall}} \\

\cmidrule(lr){2-5} \cmidrule(lr){6-9}

\textbf{Method}
& \textbf{GSM8K\textsuperscript{\(\star\)}}
& \textbf{ASDiv}
& \textbf{SVAMP}
& \textbf{Average}
& \textbf{MMLU}
& \textbf{CSQA}
& \textbf{ARC-C}
& \textbf{Average}
& \textbf{Average} \\

\midrule

\multicolumn{10}{c}{\textit{Gemma3-4B}} \\
\midrule

Base           
& 29.11 & 44.51 & 48.67 
& 40.76 
& 54.16 & 59.13 & 64.42 
& 59.24 
& 50.00 \\

$\alpha$ = 0.0                
& 48.82 & 59.09 & 51.00 & 52.97 {\color{blue}(+12.21)}
& 55.24 & 60.95 & 66.13 & 60.77 {\color{blue}(+1.53)}
& 56.87 {\color{blue}(+6.87)} \\

$\alpha$ = 0.5 \textbf{(Default)}
& 49.58 & 62.60 & 56.00 & \textbf{56.06} {\color{blue}(+15.30)}
& 55.89 & 61.10 & 67.24 & 61.41 {\color{blue}(+2.17)}
& \textbf{58.74} {\color{blue}(+8.74)} \\

$\alpha$ = 0.9
& 38.82 & 51.24 & 46.00 & 45.35 {\color{blue}(+4.59)}
& 55.59 & 60.69 & 68.00 & \textbf{61.43} {\color{blue}(+2.19)}
& 53.39 {\color{blue}(+3.39)} \\

\midrule

\multicolumn{10}{c}{\textit{Llama3.1-8B}} \\
\midrule

Base           
& 34.72 & 35.84 & 44.33 
& 38.30 
& 57.23 & 61.18 & 71.67 
& 63.36
& 50.83 \\
$\alpha$ = 0.0         
& 51.08 & 63.51 & 62.33 & 58.97 {\color{blue}(+20.67)}
& 58.81 & 60.20 & 61.95 & 60.32 {\color{red}(-3.04)}
& 59.65 {\color{blue}(+8.82)} \\

$\alpha$ = 0.5 \textbf{(Default)}      
& 51.80 & 62.70 & 66.67 & \textbf{60.39} {\color{blue}(+22.09)}
& 58.97 & 64.26 & 65.25 & 62.83 {\color{gray}(-0.53)}
& \textbf{61.61} {\color{blue}(+10.78)} \\

$\alpha$ = 0.9 
& 41.32 & 54.23 & 46.33 & 47.29 {\color{blue}(+8.99)}
& 57.34 & 68.86 & 65.78 & \textbf{63.99} {\color{gray}(+0.63)}
& 55.64 {\color{blue}(+4.81)} \\

\bottomrule
\end{tabular}%
}
\caption{\textbf{Ablation study on momentum decay $\alpha$.} Effect of varying $\alpha$ in $m_t = \alpha m_{t-1} + (1-\alpha) p_t$, evaluated on math domain for both backbones. $\alpha$ = 0.0 ignores history and oscillates; $\alpha$ = 0.9 over-relies on history and delays switching. The default $\alpha$ = 0.5 achieves the strongest overall accuracy.}
\label{table:ablation_alpha_math}
\vspace{-2mm}
\end{table*}
\begin{table*}[t]
\centering
\scriptsize
\setlength{\tabcolsep}{4pt}
\resizebox{\textwidth}{!}{%
\begin{tabular}{lccccccccc}
\toprule

\multicolumn{1}{c}{} 
& \multicolumn{4}{c}{\textbf{Medical Domain}} 
& \multicolumn{4}{c}{\textbf{General Domain}} 
& \multirow{2}{*}{\textbf{Overall}} \\
\cmidrule(lr){2-5} \cmidrule(lr){6-9}

\textbf{Method}
& \textbf{PubMedQA\textsuperscript{\(\star\)}}
& \textbf{MedQA}
& \textbf{CareQA}
& \textbf{Average}
& \textbf{MMLU}
& \textbf{CSQA}
& \textbf{ARC-C}
& \textbf{Average}
& \textbf{Average} \\
\midrule

\multicolumn{10}{c}{\textit{Gemma3-4B}} \\
\midrule

Base           
& 54.30 & 27.97 & 38.13 & 40.13 
& 54.16 & 59.13 & 64.42 & 59.24 
& 49.69 \\

$\alpha$ = 0.0                
& 66.70 & 47.29 & 43.73 & 52.57 {\color{blue}(+12.44)}
& 54.38 & 61.47 & 67.94 & 61.26 {\color{blue}(+2.02)}
& 56.92{\color{blue}(+7.23)} \\

$\alpha$ = 0.5 \textbf{(Default)}
& 66.80 
& 48.20 
& 45.19 
& \textbf{53.40} {\color{blue}(+13.27)} 
& 54.82 
& 61.26 
& 69.67 
& \textbf{61.92} {\color{blue}(+2.68)} 
& \textbf{57.66} {\color{blue}(+7.97)} \\

$\alpha$ = 0.9
& 58.60 & 39.80 & 43.56 & 47.32 {\color{blue}(+7.19)}
& 55.18 & 60.36 & 68.00 & 61.18 {\color{blue}(+1.94)}
& 54.25 {\color{blue}(+4.56)} \\

\midrule

\multicolumn{10}{c}{\textit{Llama3.1-8B}} \\
\midrule

Base           
& 68.00 & 43.99 & 50.84 & 54.28 
& 57.23 & 61.18 & 71.67 & 63.36
& 58.82 \\

$\alpha$ = 0.0         
& 76.20 & 49.96 & 55.11 & 60.42 {\color{blue}(+6.14)}
& 58.30 & 68.22 & 62.97 & 63.16 {\color{gray}(-0.20)}
& 61.79 {\color{blue}(+2.97)} \\

$\alpha$ = 0.5
\textbf{(Default)}      
& 74.60 
& 53.86 
& 56.62 
& \textbf{61.69} {\color{blue}(+7.41)}
& 58.89 
& 68.93 
& 65.27
& 64.36 {\color{blue}(+1.00)}
& \textbf{63.03} {\color{blue}(+4.21)} \\

$\alpha$ = 0.9 
& 75.80 & 49.02 & 52.18 & 59.00 {\color{blue}(+4.72)}
& 59.60 & 69.08 & 68.69 & \textbf{65.79} {\color{blue}(+2.43)}
& 62.40 {\color{blue}(+3.58)} \\

\bottomrule
\end{tabular}%
}
\caption{\textbf{Ablation study on momentum decay $\alpha$.} Same setting as Table \ref{table:ablation_alpha_math} on the medical domain.}
\label{table:ablation_alpha_medical}
\vspace{-2mm}
\end{table*}
\begin{table*}[t]
\centering
\scriptsize
\setlength{\tabcolsep}{4pt}
\resizebox{\textwidth}{!}{%
\begin{tabular}{lccccccccc}
\toprule

\multicolumn{1}{c}{} 
& \multicolumn{4}{c}{\textbf{Math Domain}} 
& \multicolumn{4}{c}{\textbf{General Domain}} 
& \multirow{2}{*}{\textbf{Overall}} \\
\cmidrule(lr){2-5} \cmidrule(lr){6-9}

\textbf{Method}
& \textbf{GSM8K\textsuperscript{\(\star\)}}
& \textbf{ASDiv}
& \textbf{SVAMP}
& \textbf{Average}
& \textbf{MMLU}
& \textbf{CSQA}
& \textbf{ARC-C}
& \textbf{Average}
& \textbf{Average} \\
\midrule

\multicolumn{10}{c}{\textit{Gemma3-4B}} \\
\midrule

Base           
& 29.11 & 44.51 & 48.67 
& 40.76 
& 54.16 & 59.13 & 64.42 
& 59.24 
& 50.00 \\

$\tau_{\mathrm{low}}$ = 0.5,
$\tau_{\mathrm{high}}$ = 0.5     
& 43.21 & 60.43 & 55.33 & 52.99 {\color{blue}(+12.23)}
& 55.36 & 61.83 & 67.06 & \textbf{61.42} {\color{blue}(+2.18)}
& 57.20 {\color{blue}(+7.20)} \\

$\tau_{\mathrm{low}}$ = 0.35,
$\tau_{\mathrm{high}}$ = 0.65   \textbf{(Default)}              
& 49.58 & 62.60 & 56.00 & 56.06 {\color{blue}(+15.30)}
& 55.89 & 61.10 & 67.24 & 61.41 {\color{blue}(+2.17)}
& \textbf{58.74} {\color{blue}(+8.74)} \\

$\tau_{\mathrm{low}}$ = 0.0,
$\tau_{\mathrm{high}}$ = 1.0    
& 49.91 & 63.64 & 56.33 & \textbf{56.63} {\color{blue}(+15.87)}
& 54.57 & 61.15 & 66.69 & 60.80 {\color{blue}(+1.56)}
& 58.72 {\color{blue}(+8.72)} \\

\midrule

\multicolumn{10}{c}{\textit{Llama3.1-8B}} \\
\midrule

Base           
& 34.72 & 35.84 & 44.33 
& 38.30 
& 57.23 & 61.18 & 71.67 
& 63.36
& 50.83 \\

$\tau_{\mathrm{low}}$ = 0.5,
$\tau_{\mathrm{high}}$ = 0.5         
& 45.51 & 54.07 & 53.33 & 50.97 {\color{blue}(+12.67)}
& 59.04 & 64.21 & 66.57 & 63.27 {\color{gray}(-0.09)}
& 57.12 {\color{blue}(+6.29)} \\

$\tau_{\mathrm{low}}$ = 0.35,
$\tau_{\mathrm{high}}$ = 0.65   \textbf{(Default)}             
& 51.80 & 62.70 & 66.67 & \textbf{60.39} {\color{blue}(+22.09)}
& 58.97 & 64.26 & 65.25 & 62.83 {\color{gray}(-0.53)}
& \textbf{61.61} {\color{blue}(+10.78)} \\

$\tau_{\mathrm{low}}$ = 0.0,
$\tau_{\mathrm{high}}$ = 1.0 
& 52.92 & 63.59 & 62.00 & 59.50 {\color{blue}(+21.20)}
& 58.15 & 65.19 & 67.83 & \textbf{63.72} {\color{gray}(+0.36)}
& \textbf{61.61} {\color{blue}(+10.78)} \\

\bottomrule
\end{tabular}%
}
\caption{\textbf{Ablation study on dispatch thresholds $(\tau_{low}, \tau_{high})$.} Effect of varying $(\tau_{low}, \tau_{high})$ in the threshold-gated 3-way dispatch, 
evaluated on math domain for both backbones. Hard switching $(0.5, 0.5)$ removes the soft-blend regime; 
fully soft $(0.0, 1.0)$ always invokes the expert, increasing latency.
The default $(0.35, 0.65)$ achieves the strongest overall accuracy.}
\label{table:ablation_threshold}
\vspace{-2mm}
\end{table*}

\subsection{Comparison with MoE-style Routing Baselines}
\label{app:moe}
We further compare CPR with intra-model MoE-style approaches that also aim to mitigate catastrophic forgetting. 
Specifically, we evaluate LoRAMoE \citep{dou2024loramoe} and MoLE \citep{wu2024mixture} under our single-domain math adaptation setting with Gemma3-4B. 
For LoRAMoE, we omit its localized balancing constraint, which is designed to balance expert groups across multiple labeled data types and is therefore not applicable to our single-domain SFT protocol. 
As shown in Table~\ref{tab:moe_comparison}, CPR achieves the highest overall average. 
While LoRAMoE improves the target-domain performance, it incurs a large degradation in general capability.
MoLE alleviates this degradation, but CPR provides a stronger balance between domain improvement and general-capability preservation.

\subsection{Generalization to Finance Domain}
\label{app:finance}
To evaluate whether CPR generalizes beyond the math and medical domains considered in our main experiments, we conduct an additional experiment on the finance domain using FinQA \citep{chen2021finqa}. 
We train the expert on the FinQA training set and evaluate on the FinQA test set as the in-domain benchmark and MMLU as an out-of-domain general benchmark. 
For FinQA, predictions are regarded as correct when they fall within a 1\% numerical relative tolerance of the gold answer under the oracle-retrieval setting.
As shown in Table~\ref{tab:finance}, SFT and EAFT substantially improve FinQA performance but suffer severe degradation on MMLU. 
In contrast, CPR retains most of the domain improvement while substantially recovering general capability, yielding the highest overall average. 
This result indicates that CPR is well generalized to finance domain as well.

\subsection{Open-ended Instruction Following}
\label{app:alpacaeval}
To evaluate CPR beyond benchmarks with deterministic target answers, we consider an open-ended instruction-following setting. 
We construct an SFT set using the top 8K examples from the Alpaca dataset \citep{alpaca}, following the same adaptation protocol as in our main experiments.
We evaluate 200 examples from AlpacaEval \citep{dubois2024length}, using pairwise comparisons against responses from the base model with Llama-3.1-70B-Instruct as the judge. 
We additionally report MMLU to assess preservation of general capability after instruction tuning.
As shown in Table~\ref{tab:alpacaeval}, CPR is preferred over the base model in 67.0\% of AlpacaEval comparisons, indicating that token-level routing remains effective for free-form generation. 
Although the SFT expert obtains a higher AlpacaEval win rate, it reduces MMLU by 6.47 points. 
CPR instead limits the MMLU drop to only 0.66 points, demonstrating a substantially more favorable balance between adaptation and general-capability preservation.

\subsection{CPR with Lightweight LoRA Expert}
\label{app:lora}
A practical limitation of CPR is that the default configuration keeps both the base and expert models resident in GPU memory. 
We therefore evaluate a lightweight variant in which the domain expert is parameterized as a LoRA adapter over the shared frozen backbone. 
This allows CPR to reuse a single copy of the backbone weights while activating the domain-specific adapter when expert computation is required.
The LoRA adapter is loaded once and toggled off for base computation and on for expert computation.
Base and expert KV caches are maintained separately, with skipped expert positions synchronized using the same batched catch-up procedure described in Section~\ref{sec:inference}.

As shown in Table~\ref{tab:lora_accuracy}, the LoRA variant retains most of CPR's accuracy gain, achieving an overall average of 56.87 compared with 58.74 for the full expert. 
More importantly, Table~\ref{tab:lora_cost} shows that its GPU memory footprint decreases from 16.60 GB to 8.83 GB, close to the 8.07 GB single-expert footprint, with only a modest latency increase from 1.40$\times$ to 1.49$\times$.

\section{Additional Qualitative Results}
\label{app:qualitative}
This section provides additional qualitative results that complement the quantitative results in Section \ref{sec4}, organized as follows: additional token-level routing examples that visualize CPR's dispatch behavior (Appendix \ref{app:token}), and a trade-off plot that empirically supports the claim that single-weight regularization-based methods operate along an inherent domain-generality trade-off (Appendix \ref{app:tradeoff}).

\subsection{Additional Token-level Routing Examples}
\label{app:token}
Figures \ref{fig:viz_csqa_math}-\ref{fig:viz_csqa_medical} provide additional token-level routing examples. Color coding follows Figure \ref{fig:viz_gsm_math}: expert tokens in red, base tokens in green, and Soft Blend tokens in intermediate shades. On in-domain queries (Figures \ref{fig:viz_gsm_math}, \ref{fig:viz_pubmed_medical}), expert calls concentrate on domain-specific tokens such as numerals and arithmetic operators in math, and biomedical terminology in medical. On out-of-domain queries (Figures \ref{fig:viz_csqa_math}, \ref{fig:viz_csqa_medical}), expert invocation is markedly sparser. These patterns provide a mechanistic view of the invocation gap in Figure \ref{fig:efficiency}: CPR preserves general capability by routing the expert precisely to tokens where domain knowledge is needed, not by uniformly damping its influence.

\subsection{Trade-off Figure}
\label{app:tradeoff}
Figure \ref{fig:tradeoff} plots the domain vs. general averages of all single-model baselines across the four settings. In every panel, a linear fit yields a negative slope, indicating that gains in domain accuracy come at the cost of general accuracy. No single-model baseline escapes this trade-off and each merely trades a different point along it. This empirically substantiates the claim in Section \ref{sec1} and motivates CPR's architectural decoupling, which steps outside the trade-off rather than relocating along it.

\section{Usage of AI Assistants}
We used AI assistants for surface-level writing support, including grammar correction, light rephrasing, and LaTeX formatting. All research ideas, experimental design, implementations, analyses, and claims were conceived and verified by the authors, who take full responsibility for the content of this paper.

\begin{table*}[t]
\centering
\scriptsize
\setlength{\tabcolsep}{4pt}
\begin{tabular}{lccccccccc}
\toprule
& \multicolumn{4}{c}{Math Domain} & \multicolumn{4}{c}{General Domain} & Overall \\
\cmidrule(lr){2-5}\cmidrule(lr){6-9}
Method & GSM8K & ASDiv & SVAMP & Avg. & MMLU & CSQA & ARC-C & Avg. & Avg. \\
\midrule
Base
& 29.11 & 44.51 & 48.67 & 40.76
& 54.16 & 59.13 & 64.42 & 59.24 & 50.00 \\

LoRAMoE
& 45.94 & 60.30 & 52.00
& 52.75 \textcolor{blue}{(+11.99)}
& 47.17 & 47.50 & 49.32
& 48.00 \textcolor{red}{(-11.24)}
& 50.38 \textcolor{gray}{(+0.38)} \\

MoLE
& 46.15 & \textbf{63.07} & \textbf{57.33}
& 55.52 \textcolor{blue}{(+14.76)}
& 53.56 & 53.40 & 61.01
& 55.99 \textcolor{red}{(-3.25)}
& 55.75 \textcolor{blue}{(+5.75)} \\

CPR (Ours)
& \textbf{49.58} & 62.60 & 56.00
& \textbf{56.06} \textcolor{blue}{(+15.30)}
& \textbf{55.89} & \textbf{61.10} & \textbf{67.24}
& \textbf{61.41} \textcolor{blue}{(+2.17)}
& \textbf{58.74} \textcolor{blue}{(+8.74)} \\
\bottomrule
\end{tabular}
\caption{\textbf{Comparison with MoE-style routing baselines.}
Test accuracy (\%) on math and general benchmarks using Gemma3-4B.}
\label{tab:moe_comparison}
\end{table*}
\begin{table*}[t]
\centering
\small
\setlength{\tabcolsep}{4pt}
\begin{tabular}{lccc}
\toprule
Method & FinQA & MMLU & Average \\
\midrule
Base
& 23.80
& \textbf{54.16}
& 38.98 \\

SFT (Expert)
& \textbf{55.80} \textcolor{blue}{(+32.00)}
& 25.26 \textcolor{red}{(-28.90)}
& 40.53 \textcolor{blue}{(+1.55)} \\

EAFT
& 53.93 \textcolor{blue}{(+30.13)}
& 26.99 \textcolor{red}{(-27.17)}
& 40.46 \textcolor{blue}{(+1.48)} \\

CPR (Ours)
& 45.90 \textcolor{blue}{(+22.10)}
& 46.87 \textcolor{red}{(-7.29)}
& \textbf{46.39} \textcolor{blue}{(+7.41)} \\
\bottomrule
\end{tabular}
\caption{\textbf{Generalization to the finance domain.}
Results with Gemma3-4B on FinQA as the in-domain benchmark and MMLU as the out-of-domain general benchmark.}
\label{tab:finance}
\end{table*}
\begin{table*}[t]
\centering
\small
\setlength{\tabcolsep}{4pt}
\begin{tabular}{lcc}
\toprule
Method & AlpacaEval Win Rate (\%) & MMLU (\%) \\
\midrule
Base
& -- & \textbf{54.16} \\

SFT (Expert)
& \textbf{76.50}
& 47.69 \textcolor{red}{(-6.47)} \\

CPR (Ours)
& 67.00
& 53.50 \textcolor{gray}{(-0.66)} \\
\bottomrule
\end{tabular}
\caption{\textbf{Open-ended instruction-following results.}
Results with Gemma3-4B on AlpacaEval, reporting the win rate against the base model, and MMLU as the out-of-domain general benchmark.}
\label{tab:alpacaeval}
\end{table*}
\begin{table*}[t]
\centering
\scriptsize
\setlength{\tabcolsep}{4pt}
\begin{tabular}{lccccccccc}
\toprule
& \multicolumn{4}{c}{Math Domain} & \multicolumn{4}{c}{General Domain} & Overall \\
\cmidrule(lr){2-5}\cmidrule(lr){6-9}
Method & GSM8K & ASDiv & SVAMP & Avg. & MMLU & CSQA & ARC-C & Avg. & Avg. \\
\midrule
Base
& 29.11 & 44.51 & 48.67 & 40.76
& 54.16 & 59.13 & 64.42 & 59.24 & 50.00 \\

CPR (Ours)
& \textbf{49.58}
& \textbf{62.60}
& \textbf{56.00}
& \textbf{56.06} \textcolor{blue}{(+15.30)}
& \textbf{55.89}
& \textbf{61.10}
& \textbf{67.24}
& \textbf{61.41} \textcolor{blue}{(+2.17)}
& \textbf{58.74} \textcolor{blue}{(+8.74)} \\

CPR (LoRA)
& 44.50
& 61.48
& 54.33
& 53.44 \textcolor{blue}{(+12.68)}
& 53.81
& 60.36
& 66.72
& 60.30 \textcolor{blue}{(+1.06)}
& 56.87 \textcolor{blue}{(+6.87)} \\
\bottomrule
\end{tabular}
\caption{\textbf{Accuracy comparison with a lightweight LoRA expert.}
Test accuracy (\%) with Gemma3-4B on math and general benchmarks, comparing CPR with the full expert and the LoRA-parameterized expert.}
\label{tab:lora_accuracy}
\end{table*}
\begin{table*}[t]
\centering
\small
\setlength{\tabcolsep}{4pt}
\begin{tabular}{lccc}
\toprule
Method & Latency (s) & Tokens/s & GPU Mem. (GB) \\
\midrule
Expert
& 5.581 (1.00$\times$)
& 17.4
& 8.07 \\

Ensemble
& 10.548 (1.89$\times$)
& 9.4
& 16.21 \\

CPR (Ours)
& 7.826 (1.40$\times$)
& 13.4
& 16.60 \\

CPR (LoRA)
& 8.298 (1.49$\times$)
& 13.3
& 8.83 \\
\bottomrule
\end{tabular}
\caption{\textbf{Runtime and memory comparison.}
Measured inference costs on GSM8K with Gemma3-4B, reporting wall-clock latency, throughput, and GPU memory usage.}
\label{tab:lora_cost}
\end{table*}
\FloatBarrier

\begin{figure*}[p]
  \centering
    \input{Figures/Viz_csqa_math}
    \vspace{40pt}
    \input{Figures/Viz_pubmedqa_medical}
    \vspace{40pt}
    \input{Figures/Viz_csqa_medical}
    \vspace{40pt}
    \input{Figures/Figure_tradeoff}
\end{figure*}
\FloatBarrier

\begin{lstfloat*}
\caption{Prompt template for GSM8K (0-shot CoT).}
\label{lst:prompt-gsm8k}
\begin{lstlisting}
Question: {question}
Answer: Let's think step by step.
\end{lstlisting}
\end{lstfloat*}

\begin{lstfloat*}
\caption{Prompt template for ASDiv (0-shot CoT).}
\label{lst:prompt-asdiv}
\begin{lstlisting}
Question: {question}
Answer: Let's think step by step.
\end{lstlisting}
\end{lstfloat*}

\begin{lstfloat*}
\caption{Prompt template for SVAMP (0-shot CoT).}
\label{lst:prompt-svamp}
\begin{lstlisting}
Question: {question}
Answer: Let's think step by step.
\end{lstlisting}
\end{lstfloat*}

\begin{lstfloat*}
\caption{Prompt template for PubMedQA (2-shot CoT). Demonstrations are drawn from MMLU and held fixed across all methods.}
\label{lst:prompt-pubmedqa}
\begin{lstlisting}
Question: Glucose is transported into the muscle cell:
(A) via protein transporters called GLUT4. (B) only in the presence of insulin. (C) via hexokinase. (D) via monocarbylic acid transporters.
Answer: Let's think step by step. Glucose (also known as the blood sugar) is the main sugar found in the human body. It is transported into the muscle cell via diffusion through protein transporters called GLUT4. The answer is (A).

Question: In a genetic test of a newborn, a rare genetic disorder is found that has X-linked recessive transmission. Which of the following statements is likely true regarding the pedigree of this disorder?
(A) All descendants on the maternal side will have the disorder. (B) Females will be approximately twice as affected as males in this family. (C) All daughters of an affected male will be affected. (D) There will be equal distribution of males and females affected.
Answer: Let's think step by step. Let's recall first that females have two X chromosomes, while males have one X and one Y chromosome. This is an important fact we need to know before answering this question. Because a male can only pass his only one X chromosome to a daughter, if he is affected by this rare genetic disorder, then we know for sure that he will pass this rare genetic disorder to all his future-born daughters. The answer is (C).

Question: {context} {question}
Options:
(A) yes
(B) no
(C) maybe
Answer: Let's think step by step.
\end{lstlisting}
\end{lstfloat*}

\begin{lstfloat*}
\caption{Prompt template for MedQA (2-shot CoT). Demonstrations are drawn from MMLU and held fixed across all methods.}
\label{lst:prompt-medqa}
\begin{lstlisting}
Question: Glucose is transported into the muscle cell:
(A) via protein transporters called GLUT4. (B) only in the presence of insulin. (C) via hexokinase. (D) via monocarbylic acid transporters.
Answer: Let's think step by step. Glucose (also known as the blood sugar) is the main sugar found in the human body. It is transported into the muscle cell via diffusion through protein transporters called GLUT4. The answer is (A).

Question: In a genetic test of a newborn, a rare genetic disorder is found that has X-linked recessive transmission. Which of the following statements is likely true regarding the pedigree of this disorder?
(A) All descendants on the maternal side will have the disorder. (B) Females will be approximately twice as affected as males in this family. (C) All daughters of an affected male will be affected. (D) There will be equal distribution of males and females affected.
Answer: Let's think step by step. Let's recall first that females have two X chromosomes, while males have one X and one Y chromosome. This is an important fact we need to know before answering this question. Because a male can only pass his only one X chromosome to a daughter, if he is affected by this rare genetic disorder, then we know for sure that he will pass this rare genetic disorder to all his future-born daughters. The answer is (C).

Question: {question}
Options:
(A) {option_A}
(B) {option_B}
(C) {option_C}
(D) {option_D}
Answer: Let's think step by step.
\end{lstlisting}
\end{lstfloat*}

\begin{lstfloat*}
\caption{Prompt template for CareQA (2-shot CoT). Demonstrations are drawn from MMLU and held fixed across all methods.}
\label{lst:prompt-careqa}
\begin{lstlisting}
Question: Glucose is transported into the muscle cell:
(A) via protein transporters called GLUT4. (B) only in the presence of insulin. (C) via hexokinase. (D) via monocarbylic acid transporters.
Answer: Let's think step by step. Glucose (also known as the blood sugar) is the main sugar found in the human body. It is transported into the muscle cell via diffusion through protein transporters called GLUT4. The answer is (A).

Question: In a genetic test of a newborn, a rare genetic disorder is found that has X-linked recessive transmission. Which of the following statements is likely true regarding the pedigree of this disorder?
(A) All descendants on the maternal side will have the disorder. (B) Females will be approximately twice as affected as males in this family. (C) All daughters of an affected male will be affected. (D) There will be equal distribution of males and females affected.
Answer: Let's think step by step. Let's recall first that females have two X chromosomes, while males have one X and one Y chromosome. This is an important fact we need to know before answering this question. Because a male can only pass his only one X chromosome to a daughter, if he is affected by this rare genetic disorder, then we know for sure that he will pass this rare genetic disorder to all his future-born daughters. The answer is (C).

Question: {question}
Options:
(A) {option_A}
(B) {option_B}
(C) {option_C}
(D) {option_D}
Answer: Let's think step by step.
\end{lstlisting}
\end{lstfloat*}

\begin{lstfloat*}
\caption{Prompt template for MMLU (2-shot CoT). Demonstrations are drawn from MMLU and held fixed across all methods.}
\label{lst:prompt-mmlu}
\begin{lstlisting}
Question: When an arguer causes confusion during refutation because of real or feigned lack of an ability to engage in refutation, that arguer may have committed the fallacy of
(A) poor sportsmanship (B) appeal to compassion (C) argument against the person (D) ignorance of refutation
Answer: Let's think step by step. Ignorance of refutation, one of Aristotle's original list of logical fallacies in his Organon, is when someone causes confusion in an argument through real or feigned inability to engage in refutation, in order to win the argument. The answer is (D).

Question: When older adults move to a new state after retirement, which of the following is the more likely destination?
(A) Texas (B) California (C) Hawaii (D) Vermont
Answer: Let's think step by step. Texas does not have state tax, and has low cost of living compared with the other three options. The answer is (A).

Question: {question}
(A) {option_A} (B) {option_B} (C) {option_C} (D) {option_D}
Answer: Let's think step by step.
\end{lstlisting}
\end{lstfloat*}

\begin{lstfloat*}
\caption{Prompt template for CommonsenseQA (2-shot CoT). Demonstrations are drawn from MMLU and held fixed across all methods.}
\label{lst:prompt-csqa}
\begin{lstlisting}
Question: When an arguer causes confusion during refutation because of real or feigned lack of an ability to engage in refutation, that arguer may have committed the fallacy of
(A) poor sportsmanship (B) appeal to compassion (C) argument against the person (D) ignorance of refutation
Answer: Let's think step by step. Ignorance of refutation, one of Aristotle's original list of logical fallacies in his Organon, is when someone causes confusion in an argument through real or feigned inability to engage in refutation, in order to win the argument. The answer is (D).

Question: When older adults move to a new state after retirement, which of the following is the more likely destination?
(A) Texas (B) California (C) Hawaii (D) Vermont
Answer: Let's think step by step. Texas does not have state tax, and has low cost of living compared with the other three options. The answer is (A).

Question: {question}
(A) {option_A} (B) {option_B} (C) {option_C} (D) {option_D} (E) {option_E}
Answer: Let's think step by step.
\end{lstlisting}
\end{lstfloat*}

\begin{lstfloat*}
\caption{Prompt template for ARC-C (2-shot CoT). Demonstrations are drawn from MMLU and held fixed across all methods.}
\label{lst:prompt-arcc}

\begin{lstlisting}
Question: When an arguer causes confusion during refutation because of real or feigned lack of an ability to engage in refutation, that arguer may have committed the fallacy of
(A) poor sportsmanship (B) appeal to compassion (C) argument against the person (D) ignorance of refutation
Answer: Let's think step by step. Ignorance of refutation, one of Aristotle's original list of logical fallacies in his Organon, is when someone causes confusion in an argument through real or feigned inability to engage in refutation, in order to win the argument. The answer is (D).

Question: When older adults move to a new state after retirement, which of the following is the more likely destination?
(A) Texas (B) California (C) Hawaii (D) Vermont
Answer: Let's think step by step. Texas does not have state tax, and has low cost of living compared with the other three options. The answer is (A).

Question: {question}
(A) {option_A} (B) {option_B} (C) {option_C} (D) {option_D}
Answer: Let's think step by step.
\end{lstlisting}
\end{lstfloat*}

\end{document}